\documentclass{article} 
\usepackage{preprint,times}

\usepackage{amsmath,amsfonts,bm}

\def\eqref#1{equation~\ref{#1}}

\def\1{\bm{1}}

\DeclareMathAlphabet{\mathsfit}{\encodingdefault}{\sfdefault}{m}{sl}
\SetMathAlphabet{\mathsfit}{bold}{\encodingdefault}{\sfdefault}{bx}{n}

\def\sA{{\mathbb{A}}}

\def\sD{{\mathbb{D}}}

\def\sL{{\mathbb{L}}}

\def\sN{{\mathbb{N}}}

\def\sQ{{\mathbb{Q}}}

\newcommand{\E}{\mathbb{E}}

\newcommand{\R}{\mathbb{R}}

\newcommand{\KL}{D_{\mathrm{KL}}}

\DeclareMathOperator*{\argmax}{arg\,max}
\DeclareMathOperator*{\argmin}{arg\,min}

\usepackage{hyperref}
\usepackage{url}

\usepackage{iftex}
\ifXeTeX
  \usepackage{fontspec}
\fi

\usepackage{amsmath,amssymb}
\usepackage{booktabs}
\usepackage{multirow}
\usepackage{array}
\usepackage{graphicx}
\usepackage{xcolor}
\usepackage{tikz}
\usetikzlibrary{arrows.meta,positioning,calc,fit,backgrounds}
\usepackage{pgfplots}
\pgfplotsset{compat=1.18}
\usepgfplotslibrary{groupplots}
\usepackage{enumitem}
\setlist{nosep,leftmargin=1.2em}
\usepackage{microtype}
\makeatletter\def\verbatim@font{\scriptsize\ttfamily}\makeatother 

\hypersetup{colorlinks=true,linkcolor=blue!60!black,citecolor=blue!60!black,urlcolor=blue!60!black}

\newcommand{\EIG}{\mathrm{EIG}}
\newcommand{\Hent}{\mathrm{H}}
\newcommand{\Th}{\Theta}
\newcommand{\hist}{h}
\newcommand{\ci}[2]{{\scriptsize$[#1,\,#2]$}}

\newcommand{\ELBOEIG}{ELBO+EIG}
\newcommand{\ELBORand}{ELBO+Random}
\newcommand{\ELBOUnc}{ELBO+Uncertainty}
\newcommand{\ExactEIG}{Exact\,Bayes+EIG}
\newcommand{\ParticleEIG}{Particle\,Bayes+EIG}
\newcommand{\LLMteacher}{LLM teacher}
\newcommand{\Student}{Llama-3.1-8B-Instruct}

\title{Learning What to Ask and What to Believe:\\Distilling a Variational Bayesian Agent\\into Language Models}

\author{Baichuan Li \\
Southern Methodist University \\
\texttt{baichuanl@smu.edu}
\And
Yanling Chang \\
Southern Methodist University \\
\texttt{ylchang@smu.edu}
\And
Alfredo Garcia \\
University of Virginia \\
\texttt{ag7s@virginia.edu}
}

\ppfinalcopy
\begin{document}

\maketitle

\begin{abstract}
Interactive tasks such as eliciting a user's preferences or diagnosing a hidden mechanism require an agent to
track its uncertainty over a latent state, ask informative questions and act on the answers. Language models
have no explicit mechanism for this, whereas Bayesian information seeking provides one through posterior
inference and expected information gain (EIG), and we ask whether it can be distilled into an 8B language
model. Our teacher, a lightweight Bayesian agent with a sequential variational belief and EIG query selection,
matches or approaches exact and particle references on three task families. We generate synthetic interactions
with the teacher on episodes disjoint from the benchmark and fine-tune \Student{} to reproduce its queries and
its belief state as a structured scratchpad. On three Multi-Turn Puzzles tasks the student raises Circuit
Decoding from 0.00 to 0.93 success, lowers the normalized rank of the recommended film from 0.50 to 0.10 and
cuts word guessing from 32 to 3.4 attempts, exceeding the strongest reported language models on the two
comparable tasks. Exposing the belief state closes much of the gap where the decision is an implicit computation rather than a read-out of the interaction; in matched teacher contrasts, EIG supervision consistently outperforms random querying, whereas greater inference fidelity does not uniformly improve the distilled student; a 70B language-model teacher does not match the Bayesian teacher in downstream student performance, and the learned skill does not transfer to an unseen task family. Variational Bayesian information seeking can thus be amortized into a language model by
transferring the belief state that drives its questions and decisions.
\end{abstract}

\section{Introduction}
\label{sec:intro}

Many useful interactions are inference problems: to recommend a film, a system must find out
what a user values; to debug a device, an engineer must probe it. In each case there is a \emph{latent state}
that the agent cannot observe but must know to act well: the weights the user places on the attributes of a
film, or the gates hidden inside the device. The agent uncovers it through a sequence of \emph{queries}, each
of which reveals a little about the state, and then commits to a \emph{decision} that is only as good as what
it has learned. Doing this well requires the two abilities that sequential Bayesian experimental design studies
together~\citep{lindley1956, chaloner1995, rainforth2024modern}: a belief over the latent state that is
updated with every answer, and the choice of the query whose answer is expected to be most informative.

Language models (LMs) do neither reliably. On the Multi-Turn Puzzles benchmark~\citep{badola2025mtp}, whose
tasks isolate this competence, the strongest reported frontier model scores only $40\%$ on the hidden circuits, and \Student{}, the open model we study, almost $0\%$; LM performance is known to degrade in multi-turn settings~\citep{laban2025lost}. Inference-time wrappers place an explicit Bayesian
computation around the LM~\citep{choudhury2025bedllm, hu2024uot, handa2024bayesian} but leave the model
unchanged and multiply the cost of every turn; training-based approaches improve the model
itself~\citep{andukuri2024stargate, zhang2024future, chen2025act, mazzaccara2024eig}, but their supervision
comes from the LM or from heuristics, so nothing guarantees that the imitated behavior is close to optimal.

\begin{figure}[t]
\centering
\resizebox{\linewidth}{!}{\definecolor{vizblue}{HTML}{2a78d6}
\colorlet{addc}{vizblue!80!black}
\begin{tikzpicture}[
  font=\scriptsize,
  box/.style={draw=black!60, rounded corners=2pt, align=center, inner sep=3pt},
  stage/.style={box, fill=black!4, text width=2.55cm},
  arr/.style={-{Stealth[length=1.8mm]}, thick, black!70},
  lab/.style={font=\tiny, text=black!70},
  ex/.style={draw=black!35, rounded corners=1.5pt, inner sep=3pt, align=left, font=\tiny\ttfamily, fill=white}
]
\node[stage] (teacher) at (0,0) {\textbf{Bayesian teacher}\\[1pt]
  belief $q_t$\\ query $=\arg\max\EIG$\\ answer $\rightarrow$ ELBO update};
\node[stage, right=0.8cm of teacher] (data) {\textbf{Synthetic episodes}\\[1pt]
  built from the public\\ task structure; disjoint\\ from the benchmark};
\node[stage, right=0.8cm of data] (student) {\textbf{Student}\\[1pt]
  \Student{}\\ + LoRA; SFT on\\ assistant tokens};
\node[stage, right=0.8cm of student] (eval) {\textbf{Interactive evaluation}\\[1pt]
  complete test splits;\\ protocol fixed\\ before each run};
\draw[arr] (teacher) -- node[above, lab] {rollouts} (data);
\draw[arr] (data) -- node[above, lab] {targets} (student);
\draw[arr] (student) -- node[above, lab] {greedy} (eval);
\node[ex, text width=4.35cm, anchor=north west] (ao) at ($(teacher.south west)+(0,-0.3)$) {%
  \textbf{\sffamily Action-only target}\\[1pt]
  {\sffamily\color{black!55}question turn}\\
  <question>Would you prefer watching The Clockwork Covenant over The Copper Tempest?</question>\\[1pt]
  {\sffamily\color{black!55}final turn}\\
  <recommend>A Song of Fate and Wind</recommend>};
\node[ex, text width=8.1cm, anchor=north west] (bs) at ($(ao.north east)+(0.35,0)$) {%
  \textbf{\sffamily Belief-scratchpad target (same turns)}\\[1pt]
  {\sffamily\color{black!55}question turn}\\
  {\color{addc}<belief>0.48, 0.42, 0.45, 0.44, 0.44, 0.45, 0.59, 0.32</belief>}\\
  <question>Would you prefer watching The Clockwork Covenant over The Copper Tempest?</question>\\[1pt]
  {\sffamily\color{black!55}final turn}\\
  {\color{addc}<belief>0.33, 0.32, 0.48, 0.65, 0.20, 0.67, 0.51, 0.49</belief>}\\
  {\color{addc}<scores>Shadows of the Hidden Valley: 3.90; The Iron Wolf's Lament: 3.97; ...; A Song of Fate and Wind: 4.18; ...</scores>}\\
  <recommend>A Song of Fate and Wind</recommend>};
\draw[arr, black!45] (data.south) -- ++(0,-0.15) -| (ao.north);
\draw[arr, black!45] (data.south) -- ++(0,-0.15) -| ($(bs.north west)+(2.5,0)$);
\end{tikzpicture}}
\caption{Overview. A Bayesian teacher interacts with synthetic episodes disjoint from the benchmark; its
interactions are serialized as \emph{actions only} or as \emph{belief scratchpad plus action} and distilled
into \Student{}, which is then evaluated interactively on the complete benchmark test splits. The targets are
shown for two turns of one movie-recommendation dialogue of the teacher; the scratchpad adds the \texttt{<belief>} and \texttt{<scores>} lines (blue; formats for all tasks in Appendix~\ref{app:scratchpads}).}
\label{fig:overview}
\end{figure}
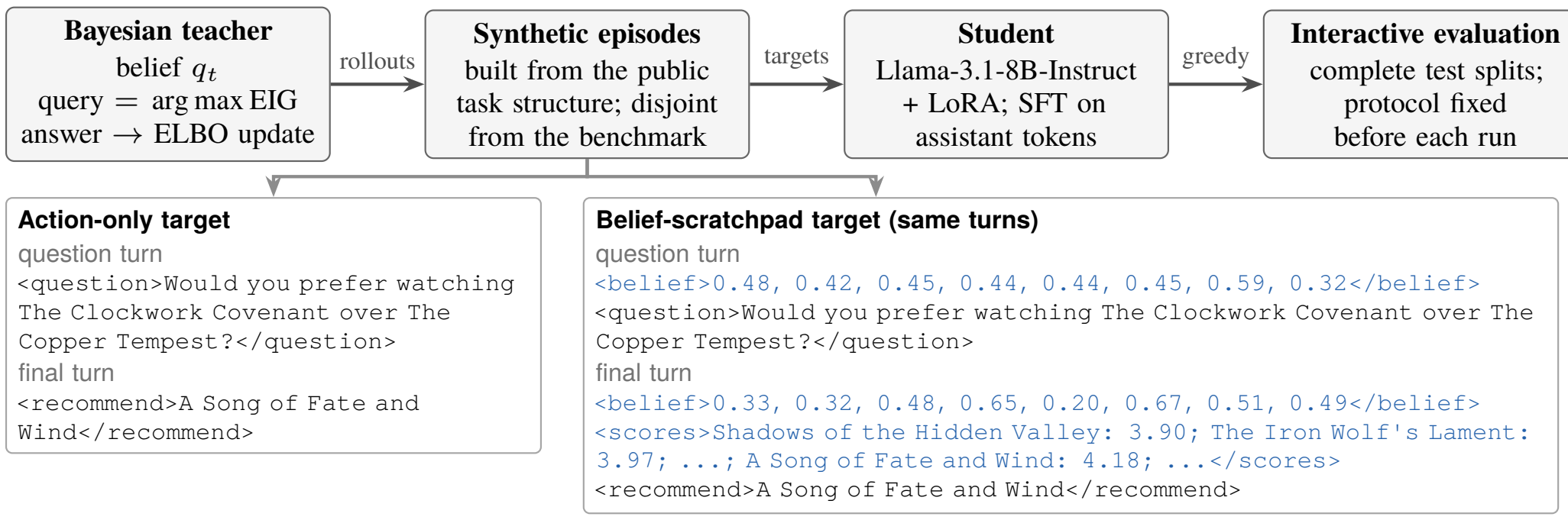

We take a third route (Figure~\ref{fig:overview}). Our teacher is a lightweight, non-neural Bayesian agent that
updates a variational belief at every turn by maximizing an evidence lower bound (ELBO) with the previous
posterior as prior, and asks the query with the highest expected information gain (EIG) under that belief.
It is cheap (Appendix~\ref{app:teacher-cost}) and verifiable against Bayesian references, and because it has an explicit belief it can supervise \emph{how} it reaches
a decision and not only \emph{what} it decides. We validate it on three task
families with continuous, structured-binary and finite-symbolic latent states, generate teacher interactions
on synthetic episodes disjoint from the benchmark, distill them into \Student{} with LoRA
fine-tuning~\citep{hu2022lora}, and evaluate on the complete test splits under a protocol fixed before each run.

Distilled from this teacher, the student closes most of the gap between the base model and the teacher. With belief-scratchpad targets, Circuit Decoding rises from 0.00 to 0.93 Success@18 (teacher 1.00), the normalized rank of the recommended film in Movie Recommendation falls from 0.496 to 0.100 (teacher 0.071), and Word Guess falls from 31.97 to 3.37 attempts (teacher 2.29). On the two tasks for which the benchmark reports comparable language-model results, the student also improves on the strongest reported model: 97.4\% versus 39.8\% circuit-wise accuracy on Circuit Decoding and a normalized rank of 0.10 versus 0.16 on Movie Recommendation.

Our contributions are threefold. First, we show that a lightweight variational Bayesian information-seeking teacher can provide effective supervision for an 8B language model on three interactive-inference tasks with continuous, structured-binary, and finite-symbolic latent states, while remaining verifiable against exact or near-exact Bayesian references. Second, we show that distilling actions alone is not always sufficient: it suffices on Circuit Decoding, where the queried bits constitute the answer and the action-only student already reaches 0.92 Success@18, but when the final decision requires inference beyond the observed interaction, action-only students gather much information yet fail to convert it into the decision, and supervising the teacher's intermediate belief and task-specific decision state substantially improves the student. Third, through matched teacher and student comparisons, we separate the effects of query selection, inference fidelity, and teacher type, finding that the query rule has the largest and most consistent effect among the Bayesian-teacher components we vary, that greater inference fidelity does not necessarily produce a better distilled student, and that a 70B language model used as teacher on the same episodes yields clearly weaker students than the Bayesian teacher. The learned competence remains task-specific: leave-one-task-out experiments show little transfer of either the information-seeking policy or its information-gain criterion to the held-out task (Section~\ref{sec:limitations}, Appendix~\ref{app:crosstask}).

\section{Problem Formulation and Background}
\label{sec:background}

\textbf{Interactive inference as sequential Bayesian experimental design.} An episode hides a latent state
$\theta^\ast\in\Th$ drawn from a known prior $p(\theta)$. At turn $t=1,\dots,T$ the agent selects a query $a_t$ from an
admissible set $\sA_t$ fixed by the task's interaction rules and observes $y_t\sim p(y\mid\theta^\ast,a_t)$; the history is
$\hist_t=(a_1,y_1,\dots,a_t,y_t)$. After the budget the agent returns a decision $d(\hist_T)$ scored against
$\theta^\ast$. An information-seeking agent therefore supplies a \emph{belief update} from $\hist_t$ to a distribution
$b_t(\theta)$, a \emph{query rule} from $b_t$ to $a_{t+1}\in\sA_{t+1}$, and a \emph{read-out} from $b_T$ to $d$. Exact
filtering gives $b_t(\theta)\propto b_{t-1}(\theta)\,p(y_t\mid\theta,a_t)$, Bayesian experimental design
gives the query rule~\citep{lindley1956, chaloner1995}, and the read-out is the Bayes decision for the task's
score. Our tasks have deterministic or near-deterministic observation models and small budgets, so every
component can be made exact or near-exact.

\textbf{Sequential variational belief updating.}\label{sec:elbo} The teacher keeps a belief $q_t$ in a fixed
variational family $\sQ$ and updates it after every observation by maximizing the per-turn evidence lower bound
with the previous belief as prior,
\begin{equation}
\label{eq:elbo}
q_t=\argmax_{q\in\sQ}\;\sL_t(q),\qquad
\sL_t(q)=\E_{q(\theta)}\big[\log p(y_t\mid\theta,a_t)\big]-\KL\big(q(\theta)\,\|\,q_{t-1}(\theta)\big),
\end{equation}
with $q_0=p(\theta)$ when the prior lies in $\sQ$ and its projection onto $\sQ$ otherwise. Up to a constant,
$\sL_t(q)=-\KL\big(q\,\|\,\tilde p_t\big)$ with $\tilde p_t\propto q_{t-1}(\theta)\,p(y_t\mid\theta,a_t)$, so
$q_t$ is the reverse-KL projection of the one-step Bayesian update onto $\sQ$, the recursion of streaming
variational Bayes~\citep{broderick2013streaming, nguyen2018vcl}: a step is exact when $\sQ$ contains
$\tilde p_t$, and the recursion recovers exact filtering when every posterior lies in $\sQ$. The streaming form
costs a constant amount per turn and yields exactly the per-turn belief that the scratchpad records; its small,
fixed parameterization is what makes the belief writable into a training target (Appendix~\ref{app:scratchpads}).
Because the ELBO requires a finite log-likelihood, the teacher's observation model for Movie Recommendation
softens the benchmark's deterministic answer rule: with $u(i)=\theta^\top\phi_i$ the utility of film $i$ under
its attribute vector $\phi_i$, the answer Yes to the pair $(i,j)$ has the Bradley--Terry probability
$\sigma\big((u(i)-u(j))/\tau\big)$, where $\sigma$ is the logistic function and $\tau=0.1$ a temperature, so that
the rule recovers the benchmark's answer for non-tied pairs as $\tau\to0$; a No Preference answer, returned only on an
exact tie, is treated as uninformative, which discards the equality it conveys (Appendix~\ref{app:teacher-movie}).
The other two tasks have deterministic observation models, which we also smooth; their categorical families
contain the exact update, so the only errors are the optimizer's and a small residual mass on refuted hypotheses.

\textbf{Expected information gain.}\label{sec:eig} Given the current belief, the teacher asks the admissible query with the
highest expected information gain,
\begin{equation}
\label{eq:eig}
a_{t+1}\in\argmax_{a\in\sA_{t+1}}\EIG_t(a),\qquad
\EIG_t(a)=\Hent\big[p_t(y\mid a)\big]-\E_{q_t(\theta)}\Hent\big[p(y\mid\theta,a)\big],
\end{equation}
where $\Hent$ is the entropy and $p_t(y\mid a)=\E_{q_t(\theta)}\,p(y\mid\theta,a)$ is the predictive under the current
belief~\citep{houlsby2011bald}, computed exactly for the categorical families and by Monte Carlo with $64$
posterior samples for the Gaussian family. Ties are broken uniformly at random under a fixed seed on Circuit
Decoding and Movie Recommendation, and on Word Guess by posterior probability and then vocabulary order
(Appendix~\ref{app:teachers}). The admissible set excludes the queries already asked on Circuit Decoding and
Word Guess, whose deterministic answers leave nothing to learn from repetition; on Movie Recommendation it is
the full set of pairs, because under the softened likelihood an asked pair keeps a positive expected gain and
may be asked again, a mismatch with the deterministic respondent that costs the turn
(Appendix~\ref{app:teacher-movie}). For a deterministic observation the second term vanishes and the rule
reduces to generalized binary search~\citep{dasgupta2004, nowak2011gbs}, and for binary feedback to
uncertainty sampling, which asks the query whose predictive probability is closest to $1/2$. The rule is
greedy: it values the next observation alone, without lookahead or regard for the final score.

\textbf{Tasks.}\label{sec:tasks} We evaluate on the three interactive-inference tasks of the Multi-Turn Puzzles
benchmark~\citep{badola2025mtp} whose latent state, query space and observation model are fully specified
(Table~\ref{tab:tasks}; Appendix~\ref{app:tasks}). They share the structure above but span different inference
regimes (a structured binary state with deterministic bits, a continuous state with graded answers, and a finite
symbolic space with multi-outcome feedback), so one teacher and one recipe are tested across latent-state and
observation structures rather than within a single task.

In \textbf{Circuit Decoding} three hidden Boolean circuits over three inputs are built from AND, OR and NOT gates
under a public inventory, $\theta$ is their joint truth table, each query reveals one of its bits, and after $18$ of
the $24$ possible queries the remaining bits are deduced from the tables being realizable under the inventory. The
teacher's family is a categorical over the tables enumerated from that inventory, the same hypothesis space and
uniform prior as the exact reference; it differs from exact Bayes only in the update, where a smoothed likelihood
and a fixed number of ELBO gradient steps leave a small residual mass on refuted tables
(Appendix~\ref{app:teacher-circuit}). The read-out is the MAP table; the benchmark scores normalized circuit-wise
accuracy, and we also report Success@18, exact recovery of all three circuits.

In \textbf{Movie Recommendation} a simulated user with hidden weights over eight film attributes scores films by a
linear utility, answers pairwise comparisons of seen films (Yes, No or No Preference), and finally receives a
recommendation from the unseen films. The family is a mean-field Gaussian on a logistic transform of the weight
range, and the read-out is the film with the highest utility under the posterior mean. The benchmark scores
$100\times(1-\text{normalized rank})$ of the recommendation; we also report Success@10, one if it is the user's
best unseen film.

In \textbf{Word Guess} one word of a small vocabulary is secret and each attempt, a vocabulary word, is answered
with per-letter feedback. The family is a categorical over the
vocabulary, a family that contains the exact posterior, so the variational and exact updates coincide. The
benchmark's normalization of the attempt count is not published, so we report raw attempts and Success@3
(Appendix~\ref{app:benchmark-metrics}).

\begin{table}[t]
\caption{The three tasks and the teacher's variational family for each ($1{,}000$ / $300$ / $400$ test episodes).}
\label{tab:tasks}
\begin{center}
\footnotesize
\setlength{\tabcolsep}{4pt}
\begin{tabular}{@{}>{\raggedright\arraybackslash}p{2.3cm}>{\raggedright\arraybackslash}p{3.6cm}>{\raggedright\arraybackslash}p{3.6cm}>{\raggedright\arraybackslash}p{3.6cm}@{}}
\toprule
 & \textbf{Movie Recommendation} & \textbf{Circuit Decoding} & \textbf{Word Guess} \\
\midrule
Latent state $\theta$ & weights $\in[0,0.9]^8$ (continuous) & $24$-bit truth table (structured binary) & one of $40$ words (finite symbolic) \\
Prior $p(\theta)$ & uniform on the grid $\{0,0.1,\dots,0.9\}^8$ & uniform over the tables realizable under the inventory & uniform over the $40$ words \\
Query $a$ & pair of seen films ($190$) & (circuit, input) ($24$) & vocabulary word ($40$) \\
Observation $y$ & Yes / No / No Preference & one bit & $5$ colors ($\le 243$ patterns) \\
Budget $T$ & $10$ & $18$ & $40$ (minimize attempts) \\
Decision $d$ & one of $40$ unseen films & all $24$ bits & the secret word \\
Benchmark metric & normalized rank of the recommended film & normalized circuit-wise accuracy & normalized number of attempts \\
Also reported & Success@10 & Success@18 & raw attempts; Success@3 \\
Teacher family $\sQ$ & mean-field Gaussian on a range transform & categorical over grammar-enumerated tables & categorical over the $40$ words \\
\bottomrule
\end{tabular}
\end{center}
\end{table}

\section{Distilling a Bayesian Teacher into a Language Model}
\label{sec:method}

\textbf{The variational information-seeking teacher.}\label{sec:teacher} The teacher, \ELBOEIG{}, combines the update of
Eq.~\ref{eq:elbo}, the rule of Eq.~\ref{eq:eig} and a task-specific read-out; the variational family and the likelihood change between tasks (Table~\ref{tab:tasks}; all hyperparameters are in Appendix~\ref{app:teachers}).
For \emph{Movie Recommendation} the belief is a mean-field Gaussian over the eight weights, mapped into the
benchmark's range, with a Bradley--Terry likelihood for the pairwise answers; the ELBO and the EIG are
estimated by Monte Carlo, and the teacher recommends the unseen film with the highest utility under the
posterior mean. For \emph{Circuit Decoding} the belief is a categorical distribution over every joint truth
table that the public gate inventory can realize, enumerated from the grammar alone; the ELBO and the EIG are
computed exactly, and the read-out is the most probable table. For \emph{Word Guess} the belief is a
categorical distribution over the vocabulary, the EIG of a guess is the entropy of the feedback partition it
induces, and the teacher commits to the most probable word once the belief has concentrated on it. Four
\emph{reference teachers} run on the same episodes: \ELBORand{} and \ELBOUnc{} keep the variational belief and
change the query rule; \ExactEIG{} (circuit, word) and \ParticleEIG{} (movie) keep the EIG rule and replace
the variational belief by exact or near-exact inference.

\textbf{Synthetic episodes and teacher interactions.}\label{sec:data} Students are never trained on benchmark episodes. For
each task a generator follows the benchmark's public structure and produces a few thousand episodes whose
latent states are disjoint from the test split, by construction and by an explicit check against the test
set. The episodes are split by latent state into training, validation and \emph{held-out synthetic} sets
(Appendix~\ref{app:data}). Every teacher is rolled out on the same episodes with the benchmark's budgets and
respondent, so the students' training data differ only in the teacher's actions and beliefs.

\textbf{Distillation targets: actions and beliefs.}\label{sec:targets} Each interaction is serialized as one chat: a fixed
system prompt, the episode's public information, alternating assistant actions and user answers, and the final
decision, with actions and decisions wrapped in tags that the evaluator parses. \emph{Action-only targets}
contain nothing but the tagged action on every assistant turn. \emph{Belief-scratchpad targets} prepend a
compact serialization of the teacher's belief, so that the student writes the belief before acting on it.
What is written follows the teacher's variational family: for Circuit Decoding, the posterior marginal of
every truth-table bit; for Movie Recommendation, the posterior mean of the weights and, before the
recommendation, the utility of every candidate film under the written weights, so that the recommendation is the argmax of a list the student has itself written; for Word Guess, the last feedback spelled out letter by
letter, a verdict for every remaining candidate, and the list of survivors (formats in
Appendix~\ref{app:scratchpads}).
Under both formats the evaluator parses the tagged action and ignores everything else, so the scratchpad is
never scored.

\textbf{Training objective and student.}\label{sec:training} The student is \Student{}~\citep{grattafiori2024llama3} with LoRA
adapters~\citep{hu2022lora}, trained by next-token prediction on the assistant tokens only. One recipe with identical hyperparameters serves every task, teacher and
target format, and the evaluated checkpoint is the one with the lowest loss on the synthetic validation split
(Appendix~\ref{app:training}).

\textbf{Controlled teacher comparisons.}\label{sec:comparisons} Episodes, recipe, student and evaluator are fixed, so a comparison between students isolates the teacher change up to training noise. We vary the \emph{query rule}
(EIG against random and uncertainty sampling, same belief), the \emph{quality of inference} (exact or particle
posterior against the variational one, same rule) and the \emph{type} of teacher, by rolling out
Llama-3.1-70B-Instruct on the same synthetic episodes as an \LLMteacher{}: its decoding is constrained to
legal actions, so its interactions are always well-formed, and its students use action-only targets and the
identical recipe.

\section{Experimental Setup}
\label{sec:setup}

\textbf{Evaluation episodes and conditions.} All benchmark results use the full released test set for each task and the same fixed evaluation protocol for all models within that task. Training, validation and held-out synthetic episodes are
generated as in Section~\ref{sec:data} and are disjoint from the test splits; no student hyperparameter or checkpoint
was chosen on the test splits, and the checks that preceded each benchmark run, against rules fixed in advance,
are listed in Appendix~\ref{app:gates}. Three choices were informed by test-split evaluations: the movie
teacher's temperature and range transform were chosen (Appendix~\ref{app:teacher-movie}), exposing beliefs followed the
failure analysis of the action-only movie students (Section~\ref{sec:res-belief}), and the character-level
word-guess scratchpad replaced a whole-word design that failed, likely because subword tokens hide the letters of a word (Appendix~\ref{app:failed-scratchpads}); the final format reproduces its benchmark results
on the held-out synthetic episodes. Table~\ref{tab:coverage} in Appendix~\ref{app:notation} lists the conditions. The design is crossed: each of the three
Bayesian teachers shared by all tasks is distilled under both target formats on all three tasks; \ELBOUnc{} is
included on Movie Recommendation, the task on which uncertainty sampling differs from EIG; and the
\LLMteacher{}, which has no belief, contributes action-only students on Circuit Decoding and Movie Recommendation. All students share one recipe (Section~\ref{sec:training}); the base model is evaluated under
the action-only prompt (Appendix~\ref{app:base-behaviour}).

\textbf{Interaction protocol.} Fixed for each condition before its benchmark run: one prompt template per
task and target format, greedy decoding with a fixed per-turn token budget, and the benchmark's deterministic
respondent. An unparsable or illegal action consumes the turn and returns an ``invalid'' message; a repeated
legal action consumes the turn and is answered consistently; a malformed decision scores zero or the worst
rank. Outputs are never repaired, re-sampled or filtered (Appendix~\ref{app:protocol}).

\textbf{Metrics.} \emph{Task success}: the benchmark's circuit-wise accuracy and Success@18 (circuit); the
benchmark's normalized-rank score and Success@10 (movie); mean attempts, with unsolved episodes counted at the
full budget, and Success@3 (Word Guess, in place of the benchmark's normalized attempt count;
Appendix~\ref{app:benchmark-metrics}).
\emph{Information-seeking quality}: the evaluator runs a \emph{reference posterior} on the model's own
history (exact for circuit and word, a near-exact particle posterior for movie), never shown to the model,
and reports the fraction of queries attaining the maximum reference EIG (\emph{EIG-optimal rate}), the mean
shortfall per query (\emph{EIG regret}; every metric is defined in Appendix~\ref{app:metric-defs}), and whether the history
identifies the latent state. \emph{Belief and read-out fidelity} for scratchpad students:
consistency of the written action with the written belief and distance of the written belief from the
reference posterior.

\textbf{Statistical analysis.} All comparisons are paired on identical episodes: the exact McNemar test on
discordant pairs~\citep{mcnemar1947} and a paired bootstrap $95\%$ confidence interval for binary outcomes,
and the Wilcoxon signed-rank test with the same bootstrap for attempt counts (Appendix~\ref{app:stats}). Each condition is one training seed with deterministic
decoding; we do not claim seed robustness.

\section{Results}
\label{sec:results}

\subsection{Teacher Validation against Exact and Particle Bayesian References}
\label{sec:res-teacher}

Table~\ref{tab:teachers} (Appendix~\ref{app:teacher-extended}) evaluates the teachers on the full released test split of each task, using the same episodes on which the students are evaluated. On Circuit Decoding, \ELBOEIG{} recovers every truth table within the query budget, matches \ExactEIG{} in mean turns to identification (11.0), and selects an exact-EIG-optimal query on 98\% of turns. On Word Guess, \ELBOEIG{} and \ExactEIG{} generate identical query sequences on all training and test episodes, because the categorical family contains the exact posterior. On Movie Recommendation, where the variational approximation is non-exact, \ELBOEIG{} approaches the particle reference on the rank-based benchmark score (92.9 vs.\ 96.5), although a gap remains in Success@10 (0.410 vs.\ 0.562). Random querying under the same belief performs worse than EIG on all three tasks. Uncertainty sampling ends slightly ahead of EIG at the teacher level on Movie Recommendation, but the two teachers are within noise on paired episodes (Appendix~\ref{app:teacher-extended}): both rules act on the same mean-field belief, and it is this approximation, not the choice between the two rules, that separates them from the particle reference. Under the reference posterior the EIG teacher's questions are in fact the more informative ones. Its one measurable handicap is that no movie teacher excludes already-asked pairs, and the EIG rule, which scores a pair by its posterior predictive, re-asks an answered pair on $6\%$ of turns, mostly among the last four questions, whereas uncertainty sampling scores pairs at the posterior mean (Eq.~\ref{eq:unc}) and moves on once an answer has shifted that mean. After distillation the ordering reverses: the \ELBOEIG{} student leads the \ELBOUnc{} student by $0.15$ in Success@10 (Table~\ref{tab:components}; Section~\ref{sec:res-components}), because the EIG policy is the one the student partly inherits, while the uncertainty student repeats most of its questions (Section~\ref{sec:res-learns}).

\subsection{Distillation Closes Most of the Gap between the Base Model and the Teacher}
\label{sec:res-main}

Table~\ref{tab:main} presents the main results. The base model never fully decodes a circuit episode,
recommends the best movie at near-chance frequency, and uses most of its query budget on Word Guess. The
distilled \ELBOEIG{} students close most of the gap between the base model and their teachers on all three
tasks. On Circuit Decoding, Success@18 rises from $0$ to $0.927$, compared with $1.00$ for the teacher. On
Movie Recommendation, the benchmark score increases from $50.4$ to $90.0$, approaching the teacher's $92.9$. On
Word Guess, mean attempts fall from $31.97$ to $3.37$, compared with $2.29$ for the teacher. These improvements
over the base model are significant under paired episode-level tests (Appendix~\ref{app:extra}).

On the two tasks with comparable published numbers, the distilled students also outperform the strongest
language model reported by the benchmark: $97.4$ versus $39.8$ in circuit-wise accuracy and $90.0$ versus
$84.3$ on the Movie Recommendation benchmark score (Appendix~\ref{app:benchmark-metrics}).

\begin{table}[t]
\caption{Main result on the complete test splits: base model, the belief-scratchpad students of the
\ELBOEIG{} teacher and of the exact or particle reference, their teachers, and the strongest language model
reported by the benchmark (Gemini~2.5~Pro; none for Word Guess: the normalization is unpublished and its two turn counts disagree). Benchmark score: circuit-wise accuracy (circuit) and
$100\times(1-\text{normalized rank})$ (movie). All student--base McNemar or Wilcoxon $p<10^{-60}$.}
\label{tab:main}
\begin{center}
\footnotesize
\setlength{\tabcolsep}{3pt}
\begin{tabular}{llcccccc}
\toprule
 & & & \multicolumn{2}{c}{\ELBOEIG{}} & \multicolumn{2}{c}{Exact / Particle Bayes+EIG} & \\
\cmidrule(lr){4-5}\cmidrule(lr){6-7}
Task & Metric & Base (8B) & Student & Teacher & Student & Teacher & Reported LM \\
\midrule
\multirow{2}{*}{Circuit} & Success@18 $\uparrow$ & $0.000$ & $\mathbf{0.927}$ & $1.000$ & $\mathbf{0.840}$ & $1.000$ & -- \\
 & benchmark score (\%) $\uparrow$ & $0.7$ & $\mathbf{97.4}$ & $100.0$ & $\mathbf{92.8}$ & $100.0$ & $39.8$ \\
\midrule
\multirow{2}{*}{Movie} & Success@10 $\uparrow$ & $0.032$ & $\mathbf{0.345}$ & $0.410$ & $\mathbf{0.409}$ & $0.562$ & -- \\
 & benchmark score (\%) $\uparrow$ & $50.4$ & $\mathbf{90.0}$ & $92.9$ & $\mathbf{93.1}$ & $96.5$ & $84.3$ \\
\midrule
\multirow{2}{*}{Word} & mean attempts $\downarrow$ & $31.97$ & $\mathbf{3.37}$ & $2.29$ & $\mathbf{3.10}$ & $2.29$ & -- \\
 & Success@3 $\uparrow$ & $0.060$ & $\mathbf{0.917}$ & $1.000$ & $\mathbf{0.917}$ & $1.000$ & -- \\
\bottomrule
\end{tabular}
\end{center}
\end{table}

\subsection{Actions Alone Are Not Always Enough: The Role of Explicit Beliefs}
\label{sec:res-belief}

Action-only supervision can transfer an effective query policy without transferring the computation that
turns the resulting interaction history into a decision. Table~\ref{tab:format} compares action-only (A) and
belief-scratchpad (B) students of the same teacher on identical test episodes. For the \ELBOEIG{} teacher on
Circuit Decoding, where the queried bits are themselves the entries of the final truth table, action-only
supervision is already sufficient: Success@18 changes only from $0.920$ to $0.927$. On the other two tasks the
action-only students are above the base model but far from their teachers, and writing the belief closes
most of that distance: Success@10 rises from $0.085$ to $0.345$ on Movie Recommendation, and mean attempts fall
from $19.2$ to $3.4$ on Word Guess.

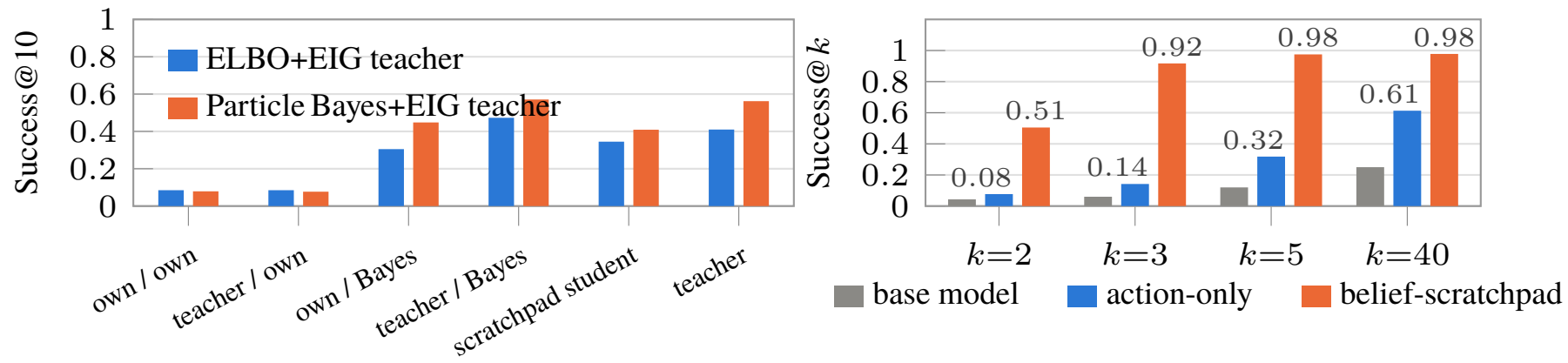
\begin{figure}[t]
\centering
\definecolor{vizblue}{HTML}{2a78d6}
\definecolor{vizorange}{HTML}{eb6834}
\definecolor{vizaqua}{HTML}{1baf7a}
\definecolor{vizgray}{HTML}{8a8985}
\begin{tikzpicture}
\begin{groupplot}[
  group style={group size=2 by 1, horizontal sep=1.0cm},
  height=3.0cm,
  ybar, /pgf/bar width=5.5pt,
  ymin=0,
  ytick={0,0.2,0.4,0.6,0.8,1.0},
  xtick pos=bottom, ytick pos=left,
  ymajorgrids, grid style={black!12}, axis line style={black!40},
  tick label style={font=\scriptsize}, label style={font=\scriptsize},
  legend style={font=\scriptsize, draw=none, fill=none, at={(0.5,-0.42)}, anchor=north, cells={anchor=west}, legend columns=-1, /tikz/every even column/.append style={column sep=0.3cm}},
  legend image code/.code={\draw[#1, draw=none] (0cm,-0.08cm) rectangle (0.22cm,0.08cm);},
  every axis plot/.append style={draw=none},
  every node near coord/.append style={font=\tiny, color=black!75, inner sep=1.5pt, /pgf/number format/fixed, /pgf/number format/precision=2},
]
\nextgroupplot[
  width=6.6cm, ymax=1.0,
  ylabel={Success@10},
  symbolic x coords={ownQ, teachQ, ownQoracle, teachQoracle, scratch, teacher},
  xtick=data,
  xticklabels={own / own, teacher / own, own / Bayes, teacher / Bayes, scratchpad student, teacher},
  xticklabel style={font=\tiny, rotate=28, anchor=north east, xshift=2pt},
  legend style={at={(0.03,0.96)}, anchor=north west, legend columns=1},
  enlarge x limits=0.1,
]
\addplot[fill=vizblue] coordinates {(ownQ,0.085) (teachQ,0.085) (ownQoracle,0.305) (teachQoracle,0.473) (scratch,0.345) (teacher,0.410)};
\addplot[fill=vizorange] coordinates {(ownQ,0.079) (teachQ,0.077) (ownQoracle,0.448) (teachQoracle,0.571) (scratch,0.409) (teacher,0.562)};
\legend{\ELBOEIG{} teacher, \ParticleEIG{} teacher}
\nextgroupplot[
  width=5.8cm, ymax=1.2,
  /pgf/bar width=6pt,
  legend style={at={(0.5,-0.30)}},
  ylabel={Success@$k$},
  symbolic x coords={k2, k3, k5, k40},
  xtick=data,
  xticklabels={$k{=}2$, $k{=}3$, $k{=}5$, $k{=}40$},
  xticklabel style={font=\scriptsize},
  enlarge x limits=0.18,
]
\addplot[fill=vizgray] coordinates {(k2,0.043) (k3,0.060) (k5,0.120) (k40,0.250)};
\addplot[fill=vizblue, nodes near coords, every node near coord/.append style={anchor=south east, xshift=4.5pt}] coordinates {(k2,0.077) (k3,0.142) (k5,0.318) (k40,0.613)};
\addplot[fill=vizorange, nodes near coords] coordinates {(k2,0.505) (k3,0.917) (k5,0.975) (k40,0.978)};
\legend{base model, action-only, belief-scratchpad}
\end{groupplot}
\end{tikzpicture}
\caption{Action-only against belief-scratchpad distillation. (a) Movie Recommendation: success of the
action-only students under their own and their teacher's questions, with their own and a Bayes-optimal
read-out, against the belief-scratchpad students and the teachers. (b) Word Guess: success within $k$
attempts for the base model and the two \ELBOEIG{} students. Labels in (a) give questions / read-out.}
\label{fig:belief}
\end{figure}

Figure~\ref{fig:belief}a diagnoses the Movie Recommendation failure by intervening separately on the questions
and on the read-out. Giving an action-only student its teacher's questions and asking only for the
recommendation leaves success nearly unchanged, whereas applying a Bayes-optimal read-out to the student's own
transcript recovers much of the teacher's success. The student's questions thus gather much of the information; what it lacks most is the read-out, a regression of the hidden weights followed by an argmax over the
candidate films (Appendix~\ref{app:movie-extra}).

Word Guess exhibits a related but distinct failure. The action-only students usually gather enough information to isolate the secret within three attempts, but then make bookkeeping errors, repeat guesses and
leave a large share of episodes unsolved (Appendix~\ref{app:word-extra}). The belief scratchpad, which makes
the surviving candidates and the checks behind them part of the training target, removes most of these
failures (Figure~\ref{fig:belief}b).

Writing the belief hurts in no final condition, although the scratchpad itself is never scored, and
it helps most where the decision requires computation beyond the recorded questions and answers. Circuit
Decoding is the cleanest exception: when the observations are the components of the answer, a strong
action-only student already approaches its teacher, and the scratchpad helps only the two weaker circuit
teachers' students, most of all the \ExactEIG{} student, whose final-answer errors it more than halves
(Appendix~\ref{app:circuit-extra}).

\begin{table}[t]
\caption{Action-only (A) against belief-scratchpad (B) students of the same teacher on identical test
episodes: score under each format and paired difference B$-$A with $95\%$ bootstrap CI.}
\label{tab:format}
\begin{center}
\scriptsize
\setlength{\tabcolsep}{1.5pt}
\resizebox{\linewidth}{!}{%
\begin{tabular}{lcccccc}
\toprule
 & \multicolumn{2}{c}{Circuit, Success@18 $\uparrow$} & \multicolumn{2}{c}{Movie, Success@10 $\uparrow$} & \multicolumn{2}{c}{Word, attempts $\downarrow$} \\
\cmidrule(lr){2-3}\cmidrule(lr){4-5}\cmidrule(lr){6-7}
Teacher & A / B & B$-$A & A / B & B$-$A & A / B & B$-$A \\
\midrule
\ELBOEIG{} & $0.920$ / $0.927$ & $+0.007$ {\tiny$[-0.037,+0.050]$} & $0.085$ / $0.345$ & $+0.260$ {\tiny$[+0.228,+0.292]$} & $19.2$ / $3.4$ & $-15.8$ {\tiny$[-17.5,-14.0]$} \\
\ELBORand{} & $0.500$ / $0.590$ & $+0.090$ {\tiny$[+0.010,+0.170]$} & $0.105$ / $0.176$ & $+0.071$ {\tiny$[+0.044,+0.098]$} & $15.2$ / $3.1$ & $-12.1$ {\tiny$[-13.4,-10.8]$} \\
\ELBOUnc{} & -- & -- & $0.064$ / $0.194$ & $+0.130$ {\tiny$[+0.103,+0.157]$} & -- & -- \\
Exact / Particle+EIG & $0.643$ / $0.840$ & $+0.197$ {\tiny$[+0.133,+0.257]$} & $0.079$ / $0.409$ & $+0.330$ {\tiny$[+0.296,+0.362]$} & $15.4$ / $3.1$ & $-12.3$ {\tiny$[-13.8,-10.8]$} \\
\bottomrule
\end{tabular}}
\end{center}
\end{table}

\subsection{Which Part of the Teacher Matters}
\label{sec:res-components}

Table~\ref{tab:components} reports the controlled contrasts of Section~\ref{sec:method} on identical episodes,
in the target format that works for each task. \textbf{Query rule.} In our matched Bayesian-teacher contrasts, replacing random querying with EIG produces a larger and more consistent change in student performance than replacing variational inference with exact or particle inference. On Word Guess the loss shows in Success@3 and not in the
mean number of attempts, because the EIG students' rare bookkeeping errors become full-budget episodes
(Section~\ref{sec:res-learns}). EIG also beats uncertainty sampling where the two rules differ, reversing the
teacher-level ordering of Table~\ref{tab:teachers}.
\textbf{Quality of inference.} Exact or particle inference with the same EIG rule helps only where the
approximation is loose. It helps on Movie Recommendation and makes no difference on Word Guess, where both
families are exact and the two teachers' training interactions coincide, so that the two students are
replicate training runs. On Circuit Decoding the effect is \emph{negative}, although the exact teacher asks optimal queries more often in training: its student imitates the teacher's queries well and fails in the
read-out after the version space has collapsed. The fidelity of a supervising model and its usefulness to the
student trained from it are thus distinct objectives, a mismatch also found for learned world models, where a more
accurate predictor need not yield a better policy trained inside it~\citep{kang2026drill}. Its validation loss is twice the variational student's
(Figure~\ref{fig:val-loss} in Appendix~\ref{app:training}), which suggests, but does not show, that the exact teacher's tie-breaking among
EIG-optimal queries yields less predictable sequences (Appendix~\ref{app:circuit-extra}). Writing the belief
removes about two thirds of this deficit but not all of it, as expected of a read-out failure. \textbf{Type of
teacher.} The \LLMteacher{} is itself weak on the synthetic episodes while producing perfectly formed
interactions, and its students inherit just that: well-formed interactions with their teacher's query
quality and almost no task success (Appendix~\ref{app:llm-teacher}). A larger general model as teacher adds
little competence; the Bayesian teacher works because it supplies informative queries \emph{and} coherent read-outs.

\begin{table}[t]
\caption{Teacher-component contrasts: paired differences between students on identical test episodes (second
condition minus first) with $95\%$ bootstrap CIs; word: Success@3 with discordant counts. Circuit uses
action-only students, movie and word belief-scratchpad students (the \LLMteacher{} row is action-only;
belief-scratchpad circuit contrasts in Appendix~\ref{app:circuit-extra}).}
\label{tab:components}
\begin{center}
\scriptsize
\setlength{\tabcolsep}{3pt}
\begin{tabular}{lccc}
\toprule
Contrast & Circuit & Movie & Word \\
\midrule
Query rule: random $\to$ EIG & $+0.420$ $[+0.350,+0.487]$ & $+0.169$ $[+0.137,+0.202]$ & $+0.205$ ($103/21$) \\
Query rule: uncertainty $\to$ EIG & identical rules & $+0.151$ $[+0.117,+0.186]$ & -- \\
Inference: variational $\to$ exact or particle & $-0.277$ $[-0.333,-0.217]$ & $+0.064$ $[+0.029,+0.098]$ & $0.000$ ($23/23$) \\
Teacher type: Bayesian $\to$ LLM teacher & $-0.920$ $[-0.950,-0.887]$ & $-0.033$ $[-0.055,-0.011]$ & -- \\
\bottomrule
\end{tabular}
\end{center}
\end{table}

\subsection{What the Student Learns}
\label{sec:res-learns}

\textbf{Students inherit their teacher's query policy.} On Circuit Decoding each student's EIG-optimal rate is
close to its teacher's, for strong and for weak teachers alike (Figure~\ref{fig:inherit} in Appendix~\ref{app:extra}), and
queries are valid and never repeated until the circuit has been identified. On the continuous task the inheritance is partial:
the students' questions attain the reference-EIG maximum far less often than their teachers', and some
students repeat questions, so question quality is the remaining gap there (Appendix~\ref{app:movie-extra}).
On Word Guess the inheritance is also partial: the students' guesses attain the exact-EIG maximum in five to six
turns in ten, against every turn for their teachers (Figure~\ref{fig:inherit}). Little is lost, because on a finite
hypothesis space a near-optimal guess is almost as informative: on solved episodes the students need about a quarter
of an attempt more than their teacher, and the secret is the only word consistent with the feedback after at most
three attempts in over $99\%$ of their episodes (Appendix~\ref{app:word-extra}).

\textbf{Written beliefs are faithful and the action follows them.} On Movie Recommendation the recommendation
is almost always the argmax of the written scores, the written belief identifies the best film about as often
as the student succeeds, and the written final belief lies close to the reference posterior mean given the
same questions. On Word Guess the elimination verdicts are correct on more than $99\%$ of turns and the
written candidate set equals the exact consistent set on most turns (Appendices~\ref{app:movie-extra}
and~\ref{app:word-extra}).

\textbf{Remaining errors.} On Circuit Decoding the \ELBOEIG{} students fail almost only after the version space
collapses to one hypothesis, mostly off by one bit: a read-out slip, not a query choice. On Movie
Recommendation the errors are mostly repeated or uninformative questions. On Word Guess they are rare
bookkeeping errors: in about $2\%$ of episodes a wrong verdict or a copying error drops the secret from the
written list, a state the teacher's traces never contain, and the student then repeats one legal guess to the
end of the budget. The random-query student has perfect bookkeeping and never dead-ends on the test split, which is why it ties
the EIG students on mean attempts there while losing on Success@3 on both the test and the held-out synthetic
episodes (Appendix~\ref{app:word-extra}).

\section{Related Work}
\label{sec:related}

\textbf{Information seeking with language models.} Language models ask suboptimal questions in
twenty-questions games~\citep{zhang2023twenty, bertolazzi2023twenty}, lose track of what they have
learned~\citep{laban2025lost}, accumulate stale or conflicting records when state is carried across many
interactions as memory unless its admission and lifecycle are explicitly governed~\citep{wang2026memguard}, and
struggle on the puzzles studied here~\citep{badola2025mtp}. Prior remedies prompt
the model to detect ambiguity and ask clarifying questions~\citep{kuhn2023clam, zhang2023clarify}, teach it to
do so from self-generated or preference data~\citep{andukuri2024stargate, zhang2024future, chen2025act}, or
optimize question selection with an information-gain reward computed by the model itself~\citep{mazzaccara2024eig,
hu2024uot}; we instead supervise with a verifiable Bayesian agent whose questions maximize EIG under its belief.

\textbf{Bayesian experimental design with language models.} BED-LLM~\citep{choudhury2025bedllm} and related
methods~\citep{handa2024bayesian, piriyakulkij2023active, kobalczyk2025disambiguation, li2023gate,
arnould2026cabed} run an EIG computation or elicitation loop around the model at test time, leaving it
unchanged and paying for it at every turn. We pay it once, offline, and amortize it into the weights, as
deep adaptive design does for neural policies~\citep{foster2021dad, ivanova2021idad, blau2022rl} and
concurrent work for a language-model policy with an EIG reward~\citep{hartmann2026amortising,
deng2026uncertainty}.

\textbf{Learning from Bayesian models and reasoning traces.} \citet{qiu2025bayesianteaching} fine-tune language models to mimic the predictions of a Bayesian model and report transfer to new recommendation and shopping domains; our setting additionally requires the student to choose what information to acquire, and our leave-one-task-out experiments do not exhibit analogous cross-task transfer. \citet{zou2025belief} show that a drifting belief corrupts credit assignment under outcome-reward reinforcement learning, and truncate the uninformative tail of such trajectories during training, on simplified variants
of two of the puzzles studied here (Appendix~\ref{app:related}); our scratchpad targets address the same
failure through supervision by an explicit belief~\citep[see also][]{li2021implicit, kim2023entity}. Distilling the outputs or rationales of a
stronger model~\citep{hinton2015distilling, ho2023reasoningteachers, hsieh2023distilling} and
scratchpads or search traces~\citep{nye2021scratchpads, wei2022cot, gandhi2024sos, lehnert2024beyondastar,
laskin2023algorithm} shows that exposing intermediate state changes what is computed or learned, and supervising the
state reached after each step rather than the final answer alone is what lets looped models track state over
sequences far longer than those seen in training~\citep{wang2026par}; we show this for a Bayesian belief in an
interactive setting (Appendix~\ref{app:related}).

\section{Limitations}
\label{sec:limitations}

\textbf{Task-specific competence.} We trained students on two task families and evaluated them zero-shot on
the third, in all three directions, under a unified interface that removes format incompatibilities. Neither the information-seeking policy nor the information-gain criterion
transfers: held-out success is within noise of the base model's, one direction shows negative transfer,
and every student solves its two source tasks (Appendix~\ref{app:crosstask}); what is learned is a task-specific procedure.

\textbf{Scope.} The teacher needs an explicit hypothesis space and likelihood. We use one student backbone,
one training seed per condition, greedy decoding and a deterministic respondent. The single replicate pair available (two
word-guess students whose training data coincide) agrees closely when the students work and differs by
several attempts when they fail (Appendix~\ref{app:word-extra}). Question quality on the continuous task remains
below the teacher's, and about two percent of word-guess episodes end in an unrecoverable bookkeeping error,
a state the teacher's traces never contain. The written belief carries no confidence of its own, so a corrupted
state is neither flagged nor repaired; persisting confidence and uncertainty descriptors alongside the state, as
recent agent-memory governance does~\citep{wang2026memguard}, is one possible remedy.

\section{Conclusion}
\label{sec:conclusion}

A lightweight Bayesian agent that updates a variational belief and asks the most informative query is
close to exact or particle references on three tasks where language models struggle, and can be distilled
into an 8B model from synthetic interactions alone. The recipe works when the target exposes what the teacher
knows, not only what it does: where the decision must be computed, a written belief closes most of the
gap. Among the teacher components we vary, the information-gain rule has the largest and most consistent
effect, whereas greater inference fidelity does not necessarily improve distillation; a 70B language-model
teacher yields weaker students than the Bayesian teacher.

\subsection*{AI use statement}
In this work, we used generative AI tools (large language models used as coding and writing assistants) to
implement methods, namely the teachers, the episode generators, the serialization of teacher traces into
training targets, and the training and evaluation code, and to provide feedback on the experimental
methodology and to draft interpretations of results, in both cases from analyses that the authors specified
and reviewed. Apart from the \LLMteacher{} baseline of Appendix~\ref{app:llm-teacher}, in which
Llama-3.1-70B-Instruct is rolled out on the same synthetic episodes as an experimental comparison and its
action-only dialogues train the corresponding students, we have not used generative AI tools to generate data
sets: every synthetic episode is produced by the deterministic generators of Appendix~\ref{app:generators}. We
have not used generative AI tools to develop the theoretical models or conceptual framework, to propose or
refine hypotheses, to design the experiments, or to formulate or prove mathematical claims. Translation,
data-set cleaning and qualitative or thematic data analysis are not applicable to this work. Additionally, we
used generative AI tools to identify and summarize related literature, to draft and edit parts of the
manuscript for readability, to create and modify tables and figures from the experiments' output files, to run
and monitor experiments, and to format references. We have reviewed all AI-assisted work: all experimental
designs, decision rules and interpretations were specified or reviewed by the authors; AI-generated code was
reviewed and tested by the authors, with the integrity checks, smoke runs and pre-registered validation gates
described in Appendix~\ref{app:gates}; every experimental number in the paper was traced back to the output
files of the evaluator, teacher or training run that produced it, every language-model score quoted from the
benchmark to the benchmark paper, and every citation surfaced with AI assistance was verified against the
original source. We take responsibility for the final content of this work, including text, claims and
artifacts produced with the aid of generative AI.

\subsection*{Ethics statement}
This work uses synthetic interactive puzzles with simulated respondents and publicly released models and
benchmark data; no human subjects or personal data are involved. Improved information-seeking ability in
language models could be misused to elicit information from people more effectively; the tasks studied here
are fully specified puzzles with a known latent state, and the learned competence did not transfer across the
three task families (Appendix~\ref{app:crosstask}), which limits this concern for the present work.

\subsection*{Reproducibility statement}
Every benchmark result in the main text is computed on the complete released test split of the corresponding
task under an evaluator whose prompts, decoding settings, parsing rules and scoring are specified in
Appendix~\ref{app:protocol}, with hashes of prompts, adapters and episode manifests verified inside every run;
results on training, validation or held-out synthetic episodes, on a subset of a test split, or under the
unified interface of Appendix~\ref{app:crosstask} are labeled as such where they appear.
The teachers are specified with all hyperparameters in Appendix~\ref{app:teachers}; the generators, splits,
serialization formats and training configuration in Appendix~\ref{app:data}; the statistical procedures in
Appendix~\ref{app:stats}; software versions, hardware, runtimes and seeds in Appendix~\ref{app:repro}; and the
prompts in Appendix~\ref{app:prompts}. No code or data accompany this submission; the code, synthetic
episode banks, teacher traces, training data and LoRA adapters for every condition will be released upon
acceptance (Appendix~\ref{app:repro}).

\bibliography{references}
\bibliographystyle{preprint}

\clearpage
\appendix
\section{Notation and Summary of Conditions}
\label{app:notation}

\subsection{Notation}
\begin{center}\footnotesize
\begin{tabular}{lp{10.5cm}}
\toprule
$\theta,\ \theta^\ast,\ \Th$ & latent state, its true value, the hypothesis space \\
$a_t\in\sA_t,\ y_t$ & query at turn $t$ from the admissible set $\sA_t$ (unasked queries on circuit and word; all pairs on movie), its observation \\
$\hist_t$ & history $(a_1,y_1,\dots,a_t,y_t)$; $T$ the query budget \\
$d(\hist_T)\in\sD$ & terminal decision (recommendation, truth table, secret word) \\
$p(y\mid\theta,a)$ & observation model of the task (deterministic for circuit and word) \\
$b_t$ & exact Bayesian posterior given $\hist_t$; the \emph{reference posterior} of the evaluator \\
$q_t\in\sQ$ & the teacher's variational belief after turn $t$; $\sQ$ its family \\
$\sL_t(q)$ & per-turn evidence lower bound, Eq.~\ref{eq:elbo}, with $\beta=1$ \\
$\EIG_t(a)$ & expected information gain of query $a$ under $q_t$, Eq.~\ref{eq:eig} \\
$\varepsilon$ & smoothing of a deterministic likelihood ($0.01$ circuit, $10^{-6}$ word) \\
$\phi_i$ & attribute vector of film $i$ (Movie Recommendation) \\
$\mathrm{pred}[k,a]$ & answer predicted by hypothesis $\theta_k$ to query $a$ (Circuit Decoding) \\
\bottomrule
\end{tabular}
\end{center}

\subsection{Glossary of Teachers, Students and Target Formats}
\label{app:glossary}
\begin{center}\small
\begin{tabular}{>{\raggedright\arraybackslash}p{3.4cm}p{9.6cm}}
\toprule
\ELBOEIG{} & the proposed teacher: sequential variational belief (Section~\ref{sec:elbo}), EIG query rule (Section~\ref{sec:eig}), task read-out \\
\ELBORand{} & same belief, uniformly random admissible query \\
\ELBOUnc{} & same belief, the pair whose predicted answer is closest to even odds (movie only; equals EIG on circuit) \\
\ExactEIG{} & exact posterior by hard filtering of hypotheses inconsistent with the observations; exact EIG (circuit, word) \\
\ParticleEIG{} & $200{,}000$-particle posterior on the benchmark's weight grid, exact reweighting, closed-form EIG (movie) \\
\LLMteacher{} & Llama-3.1-70B-Instruct, greedy decoding under a legality constraint, on the same synthetic episodes \\
Base model & \Student{} without adapters, under the action-only prompt \\
Action-only student & \Student{} + LoRA trained on action-only targets (Section~\ref{sec:targets}) \\
Belief-scratchpad student & \Student{} + LoRA trained on belief-scratchpad targets \\
Complete test split & all released test episodes of a task ($300$ / $1{,}000$ / $400$) \\
Held-out synthetic episodes & generator episodes disjoint from training, validation and the benchmark ($464$ / $400$ / $400$) \\
Reference posterior & exact (or particle) posterior computed by the evaluator on the model's own history; never shown to the model \\
Leave-one-task-out student & one student trained on two task families and evaluated zero-shot on the third (Appendix~\ref{app:crosstask}) \\
\bottomrule
\end{tabular}
\end{center}

\begin{table}[htbp]
\caption{Conditions evaluated on the complete test splits. A = action-only targets, B = belief-scratchpad
targets. Uncertainty sampling coincides with EIG on Circuit Decoding; the \LLMteacher{} has no belief and
therefore no scratchpad condition.}
\label{tab:coverage}
\begin{center}
\footnotesize
\setlength{\tabcolsep}{4pt}
\begin{tabular}{lccc}
\toprule
Condition & Circuit Decoding & Movie Recommendation & Word Guess \\
\midrule
Base model & yes & yes & yes \\
\ELBOEIG{} student & A, B & A, B & A, B \\
\ELBORand{} student & A, B & A, B & A, B \\
\ELBOUnc{} student & (= EIG) & A, B & -- \\
Exact / Particle Bayes+EIG student & A, B & A, B & A, B \\
\LLMteacher{} student & A & A & -- \\
\bottomrule
\end{tabular}
\end{center}
\end{table}

\section{Task Specifications}
\label{app:tasks}
Every test split in this paper is the \texttt{test} split of the benchmark's public \mbox{Hugging Face} release
(\path{arianhosseini/mt_puzzles}; subsets \path{circuit_decoding}, \path{movie_recommendation} and
\path{word_guess})~\citep{badola2025mtp}, downloaded once and frozen by per-row hashes in a manifest that
every evaluator verifies. All teacher and student numbers reported on a test split are computed on these
released episodes; the episodes we generate (Appendix~\ref{app:data}) are used only for training,
validation and the held-out checks of Appendix~\ref{app:gates}.

\subsection{Movie Recommendation}
Each episode provides an ordered list of eight attribute names, $20$ seen films and $40$ unseen films, each
with an attribute vector $\phi\in\R^8$ (values with one or two decimals, sparse in three attributes), and a
hidden weight vector $\theta^\ast\in\{0,0.1,\dots,0.9\}^8$. The utility of a film is $\theta^{\ast\top}\phi$.
A question names two seen films $(A,B)$; the deterministic user answers Yes if $u(A)>u(B)$, No if $u(A)<u(B)$
and No Preference on an exact tie (utilities are exact multiples of $10^{-3}$; ties occur in $1{,}134$ of the
$190{,}000$ seen pairs of the test split). Ten questions, then one recommendation from the unseen list.
Success@10 is $1$ if the recommended film attains the maximum unseen utility ($7$ test episodes have a tied
optimum, and any optimal film counts). The test split has $1{,}000$ episodes: $20$ score tables, each served
to $50$ users.

\subsection{Circuit Decoding}
Three circuits $A$, $B$, $C$ over inputs $(x_0,x_1,x_2)$ are built from AND, OR and NOT gates under a public
total inventory; the test split contains three inventories, AND/OR/NOT $=3/3/0$, $3/3/1$ and $1/5/2$, with
$100$ episodes each. The latent state is the joint $24$-bit table (bits $1$--$8$ circuit $A$ on inputs
$000,\dots,111$; bits $9$--$16$ circuit $B$; bits $17$--$24$ circuit $C$). A query
\texttt{<query>C(x0,x1,x2)</query>} is answered with the bit. Eighteen queries, then the $24$-bit
\texttt{<guess>}. Success@18 is exact match; we also report bitwise accuracy.

\subsection{Word Guess}
Each episode gives a vocabulary of $40$ five-letter words, of which one is secret; the test split has $10$
vocabularies, each word being the secret exactly once ($400$ episodes). An \texttt{<attempt>} must be a
vocabulary word and is answered with five colors computed by the two-pass rule (greens first, then yellows
for remaining letters, duplicates handled by count). The episode ends at the correct guess or after $40$
attempts. The score is the number of attempts (unsolved $=40$); Success@$k$ is the fraction solved within $k$.

\subsection{Interaction Rules and Scoring}
\label{app:rules}
Identical for teachers and students: an illegal or unparsable action consumes the
turn and is answered ``Invalid query.'' (or ``Invalid question.'' / invalid attempt) with no information; a
repeated legal action consumes the turn and is answered consistently; a malformed terminal decision scores
$0$ (circuit), the worst rank (movie) or an unsolved episode (word).

\subsection{Correspondence with the Metrics Reported by the Benchmark}
\label{app:benchmark-metrics}
The benchmark scores Circuit Decoding by a normalized \emph{circuit-wise accuracy}, which we take to be the
fraction of individual circuits whose truth table is recovered exactly. We report it as the benchmark score
next to Success@18, the stricter criterion that all three circuits of an episode are recovered, which is a
lower bound on it. On the test split the circuit-wise accuracy is $0.7\%$ for the base model, $97.0\%$,
$80.7\%$ and $86.8\%$ for the action-only \ELBOEIG{}, \ELBORand{} and \ExactEIG{} students, $97.4\%$, $84.4\%$
and $92.8\%$ for their belief-scratchpad counterparts, $2.6\%$ for the
\LLMteacher{} student and $100.0\%$ for the two EIG teachers. For Movie Recommendation the benchmark reports the
normalized rank of the recommended film, shown as a success rate $100\times(1-\text{normalized rank})$ with
normalized rank $=(\text{rank}-1)/39$ (ties share the best rank; $0.5$ is a random pick). For Word Guess it
reports a normalized attempt count whose scorer is not published, and the two mean turn counts it gives for
Gemini~2.5~Pro on this task disagree (its Table~2 lists $2.4$ average turns; its Section~3.1 text states
$5.37$), so we report raw attempts and make no language-model comparison on this task; the corresponding
cell of Table~\ref{tab:main} is left empty. The
language-model values for the other two tasks are read from the summary figure of the benchmark
paper~\citep{badola2025mtp}: Gemini~2.5~Pro, Claude~3.5~Sonnet and GPT-4o score $39.8$, $9.3$ and $15.4$ on
Circuit Decoding and $84.3$, $82.0$ and $62.4$ on Movie Recommendation. All of these published values were
obtained under the benchmark's own prompts and each API's default sampling parameters (Section~3 of the
benchmark paper), not under the evaluator of Appendix~\ref{app:protocol} (our prompt templates, greedy decoding,
fixed token budgets), so their comparison with our students (Table~\ref{tab:main}, Section~\ref{sec:res-main})
is indicative rather than paired.

\section{The Bayesian Teachers}
\label{app:teachers}

\subsection{The Sequential Variational Update and Its Relation to Exact Filtering}
\label{app:elbo-derivation}
For a single observation $y_t$ the log marginal likelihood under the current prior $q_{t-1}$ satisfies
$\log\int q_{t-1}(\theta)p(y_t\mid\theta,a_t)\,d\theta\ \ge\ \sL_t(q)$ for every $q$, with equality if and only
if $q\propto q_{t-1}(\theta)\,p(y_t\mid\theta,a_t)$, that is, at the exact Bayesian update. Hence maximizing
$\sL_t$ over an unrestricted family recovers exact filtering, and maximizing it over a restricted family $\sQ$
returns the member of $\sQ$ that minimizes $\KL(q\,\|\,\tilde b_t)$ where $\tilde b_t$ is the exact update of
the previous variational belief. Freezing $q_t$ and using it as the prior of turn $t+1$ gives the recursion of
streaming variational Bayes~\citep{broderick2013streaming}; the approximation error accumulates only through
the projections. For a categorical $\sQ$ over a finite hypothesis space the projection is exact, and the remaining
errors are those of the finite optimizer and of the likelihood smoothing $\varepsilon$ (Appendices~\ref{app:teacher-circuit} and~\ref{app:teacher-word}).

\subsection{Movie Recommendation}
\label{app:teacher-movie}
Family: $q_t(z)=\sN(\mu_t,\mathrm{diag}\,s_t^2)$ on $z\in\R^8$ with $\theta=0.9\,\sigma(z)$, so that
$\theta\in(0,0.9)^8$ matches the benchmark's range; initial $\mu_0=0$, $s_0=1$. Likelihood: the
Bradley--Terry model $p(\text{Yes}\mid\theta,(i,j))=\sigma(\theta^\top(\phi_i-\phi_j)/\tau)$ with $\tau=0.1$;
a No Preference answer is treated as uninformative for the update. Update: the ELBO expectation is estimated
with $16$ reparameterized samples and maximized with Adam over $(\mu,\log s)$; EIG: Monte Carlo over $64$
posterior samples for every one of the $190$ pairs (asked pairs are not excluded; a re-asked pair is answered
consistently and consumes the turn), ties broken uniformly. Read-out: the unseen film with the highest
utility under $\theta(\mu_T)$. The range transform and the temperature raised the teacher's test-split
Success@10 from $0.29$ to $0.42$ during development on a Bradley--Terry respondent; nothing was tuned on the
deterministic-respondent protocol used for all reported numbers.

\subsection{Circuit Decoding}
\label{app:teacher-circuit}
Hypothesis space: all triples of circuits realizable with the public inventory, enumerated from the grammar
(every allocation of gates to the three circuits, every tree shape and leaf order, every NOT placement),
de-duplicated on the joint table: $K=252$, $3{,}924$ and $2{,}886$ tables for the three inventories (prior
entropies $8.0$, $11.9$ and $11.5$ bits); the grammar was verified to reproduce all $900$ circuits of the test
split. Prediction matrix $\mathrm{pred}\in\{0,1\}^{K\times24}$. Family: $q_t=\mathrm{softmax}(\lambda_t)$,
$\lambda_t\in\R^K$. Likelihood: $p_\varepsilon(y\mid\theta_k,a)=1-\varepsilon$ if $\mathrm{pred}[k,a]=y$, else
$\varepsilon$, with $\varepsilon=0.01$. Update: exact $K$-term ELBO, Adam (learning rate $0.05$, $80$ steps)
warm-started at $\lambda=\log q_{t-1}$, $\beta=1$. EIG:
$p_t(y{=}1\mid a)=\sum_k q_t(\theta_k)[(1-\varepsilon)\mathrm{pred}[k,a]+\varepsilon(1-\mathrm{pred}[k,a])]$
and $\EIG_t(a)=\Hent_2[p_t(y{=}1\mid a)]-\Hent_2(\varepsilon)$ over unasked queries, ties uniform. Read-out:
the maximum-a-posteriori table. Because every hypothesis consistent with the data receives the same update,
$q_t$ is exactly uniform on the version space; the only error is mass left on refuted hypotheses (total
variation $0.027$ from the exact posterior at turn $18$), and the MAP table is always a consistent hypothesis.

\subsection{Word Guess}
\label{app:teacher-word}
Family: categorical over the $40$ words, uniform prior. Likelihood: the deterministic coloring
$f(a,\theta)$ smoothed as $p_\varepsilon(y\mid\theta,a)=1-\varepsilon$ if $y=f(a,\theta)$, else
$\varepsilon/242$, with $\varepsilon=10^{-6}$. Update: exact expectation over the $40$ hypotheses, Adam on the
logits ($1{,}600$ steps, learning rate $0.5$), which brings every update within $\KL<10^{-4}$ of the analytic
posterior. EIG: entropy of the predictive feedback partition of a candidate guess (smoothed variant), over all
unguessed words; ties within $10^{-4}$ broken by posterior probability, then vocabulary order. Commit rule:
guess the maximum-a-posteriori word once its belief exceeds $0.999$. The policies receive only the vocabulary,
the belief and the feedback matrix; the first guess is therefore a fixed function of the vocabulary.

\subsection{Reference Teachers}
\label{app:teacher-refs}
\ELBORand{}: the same belief; the query is drawn uniformly from the admissible set $\sA_t$ (five paired seeds per
episode). \ELBOUnc{} (movie): uncertainty sampling under the same belief, which asks the pair whose predicted
answer under the posterior-mean weights is closest to even odds,
\begin{equation}
\label{eq:unc}
a_{t+1}=\argmin_{(i,j)\in\sA}\big|\hat p_t(\text{Yes}\mid i,j)-\tfrac12\big|,\qquad
\hat p_t(\text{Yes}\mid i,j)=\sigma\big(\bar\theta_t^{\top}(\phi_i-\phi_j)/\tau\big),\quad
\bar\theta_t=\E_{q_t}[\theta],
\end{equation}
that is, the pair with the smallest $|\bar\theta_t^{\top}(\phi_i-\phi_j)|$; unlike Eq.~\ref{eq:eig} it uses the mean
of the belief and not its spread. Asked pairs are not excluded by any movie teacher (Appendix~\ref{app:teacher-extended}).
On Circuit Decoding, where every answer is one bit, uncertainty sampling coincides with Eq.~\ref{eq:eig}. \ExactEIG{}: hard
filtering, $w_t(\theta_k)\propto w_{t-1}(\theta_k)\mathbf 1[\mathrm{pred}[k,a_t]=y_t]$, and exact EIG
$\Hent_2[\sum_k w_t(\theta_k)\mathrm{pred}[k,a]]$ (circuit) or the exact partition entropy (word).
\ParticleEIG{} (movie): $200{,}000$ particles on the grid $\{0,\dots,0.9\}^8$, exact reweighting under the
deterministic answer model with a temperature-aware likelihood, closed-form EIG; this posterior also serves as
the evaluator's reference posterior for Movie Recommendation.

\subsection{Hyperparameters and Computational Cost}
\label{app:teacher-cost}
\begin{center}\footnotesize
\setlength{\tabcolsep}{4pt}
\begin{tabular}{>{\raggedright\arraybackslash}p{3.3cm}>{\raggedright\arraybackslash}p{3.1cm}>{\raggedright\arraybackslash}p{3.1cm}>{\raggedright\arraybackslash}p{3.1cm}}
\toprule
 & Movie & Circuit & Word \\
\midrule
Family & Gaussian, $16$ parameters & categorical, $K\le3{,}924$ & categorical, $40$ \\
Optimizer & Adam, $16$ ELBO samples & Adam, lr $0.05$, $80$ steps & Adam, lr $0.5$, $1{,}600$ steps \\
Likelihood smoothing & $\tau=0.1$ & $\varepsilon=0.01$ & $\varepsilon=10^{-6}$ \\
EIG estimate & $64$ samples & exact & exact (smoothed) \\
Wall time per episode (CPU) & $3.3$ s & $0.25$ s & $<0.7$ s \\
Reference teacher per episode & $54$ s (particle) & $0.003$ s (exact) & $<0.1$ s (exact) \\
\bottomrule
\end{tabular}
\end{center}

\subsection{Extended Teacher Results}
\label{app:teacher-extended}
\begin{table}[htbp]
\caption{Teacher validation on the complete released test splits ($300$ / $1{,}000$ / $400$ episodes; five
paired seeds on Circuit Decoding and for \ELBORand{} on Word Guess, one run on Movie Recommendation).
EIG-optimal rate and regret are measured under the exact posterior. \ELBOUnc{} asks the pair whose predicted
answer is closest to even odds (Eq.~\ref{eq:unc}, Appendix~\ref{app:teacher-refs}) and is run only on Movie
Recommendation, where it differs from EIG.}
\label{tab:teachers}
\begin{center}
\footnotesize
\setlength{\tabcolsep}{2.5pt}
\begin{tabular}{llcccc}
\toprule
Task & Metric & ELBO+EIG & ELBO+Rand. & ELBO+Unc. & Exact / Part. \\
\midrule
\multirow{3}{*}{Circuit} & Success@18 $\uparrow$ & $\mathbf{1.00}$ & $0.70$ & -- & $\mathbf{1.00}$ \\
 & turns to identify $\downarrow$ & $11.0$ & $17.1$ & -- & $11.0$ \\
 & EIG-opt. / regret (bits) & $0.98$ / $0.001$ & $0.30$ / $0.33$ & -- & $1.00$ / $0$ \\
\midrule
\multirow{2}{*}{Movie} & Success@10 $\uparrow$ & $0.410$ & $0.373$ & $0.428$ & $\mathbf{0.562}$ \\
 & benchmark score (\%) $\uparrow$ & $92.9$ & $91.0$ & $93.8$ & $\mathbf{96.5}$ \\
\midrule
\multirow{2}{*}{Word} & mean attempts $\downarrow$ & $\mathbf{2.29}$ & $3.09$ & -- & $\mathbf{2.29}$ \\
 & Success@3 $\uparrow$ & $\mathbf{1.00}$ & $0.73$ & -- & $\mathbf{1.00}$ \\
\bottomrule
\end{tabular}
\end{center}
\end{table}
\textbf{Circuit Decoding, per inventory} ($300$ test episodes $\times$ $5$ paired seeds).
\begin{center}\scriptsize
\setlength{\tabcolsep}{3pt}
\begin{tabular}{lccc|ccc}
\toprule
& \multicolumn{3}{c|}{Success@12} & \multicolumn{3}{c}{queries to identification} \\
Inventory ($K$) & ELBO+EIG & ELBO+Rand. & Exact+EIG & ELBO+EIG & ELBO+Rand. & Exact+EIG \\
\midrule
3/3/0 (252)  & $0.88$ & $0.32$ & $0.87$ & $9.3$  & $16.3$ & $9.2$ \\
3/3/1 (3924) & $0.63$ & $0.18$ & $0.58$ & $12.3$ & $17.8$ & $12.3$ \\
1/5/2 (2886) & $0.82$ & $0.25$ & $0.84$ & $11.5$ & $17.3$ & $11.3$ \\
\bottomrule
\end{tabular}
\end{center}
Overall: Success@18 $1.00$ / $0.70$ / $1.00$; bitwise accuracy at $18$: $1.000$ / $0.972$ / $1.000$;
posterior mass on $\theta^\ast$ at $18$: $0.97$ / $0.62$ / $1.00$; total variation to the exact posterior at
$18$: $0.027$ / $0.123$ / $0$; exact-EIG regret $0.001$ / $0.329$ / $0$ bits per query.

\textbf{Movie Recommendation} ($1{,}000$ test episodes, deterministic respondent).
\begin{center}\footnotesize
\setlength{\tabcolsep}{3pt}
\begin{tabular}{lccccc}
\toprule
Teacher & Success@10 & norm.\ rank & benchmark \% & rank of true best & No Pref.\ rate \\
\midrule
\ELBOEIG{} & $0.410\pm0.016$ & $0.071$ & $92.9$ & $3.40$ & $0.011$ \\
\ELBORand{} & $0.373\pm0.015$ & $0.090$ & $91.0$ & $3.88$ & $0.006$ \\
\ELBOUnc{} & $0.428\pm0.016$ & $0.062$ & $93.8$ & $3.07$ & $0.014$ \\
\ParticleEIG{} & $0.562\pm0.016$ & $0.035$ & $96.5$ & $2.38$ & $0.031$ \\
\bottomrule
\end{tabular}
\end{center}
On the synthetic training episodes the same teachers reach $0.434$ / $0.385$ / $0.444$ / $0.546$, with
reference-EIG-optimal rates $0.142$ / $0.005$ / $0.015$ / $0.841$ and reference-EIG regret $0.055$ / $0.203$ /
$0.122$ / $0.001$ nats per question. Paired on the $1{,}000$ test episodes, \ELBOUnc{} minus \ELBOEIG{} in
Success@10 is $+0.018$ ($191$ / $173$ discordant episodes, exact McNemar $p=0.37$, $95\%$ bootstrap CI
$[-0.018,+0.055]$); the two teachers' per-turn success curves stay within $0.02$ of each other at every turn, and
their question sets are nearly disjoint (mean Jaccard overlap $0.035$). No movie teacher excludes asked pairs: the
repeated-question rates are $0.064$ / $0.022$ / $0.014$ / $0.024$ (\ELBOEIG{} / \ELBORand{} / \ELBOUnc{} /
\ParticleEIG{}), and $84\%$ of the \ELBOEIG{} repeats fall in the last four questions.

\textbf{Word Guess} ($400$ test episodes). Mean attempts $2.29\pm0.03$ / $3.09\pm0.03$ / $2.29\pm0.03$;
Success@1/2/3: $0.025$/$0.682$/$1.000$ for both EIG teachers, $0.024$/$0.317$/$0.733$ for \ELBORand{};
Success@5 $1.000$ / $0.967$ / $1.000$. Paired: \ELBOEIG{} saves $0.79$ attempts against \ELBORand{} ($95\%$
bootstrap CI $[0.72,0.87]$, cluster CI over vocabularies $[0.74,0.85]$, Wilcoxon $p=5.7\times10^{-52}$; better
on $328$, worse on $47$, tied on $25$ episodes) and produces guess sequences identical to \ExactEIG{} on all
$400$ episodes.

\begin{figure}[t]
\centering
\includegraphics[width=\linewidth]{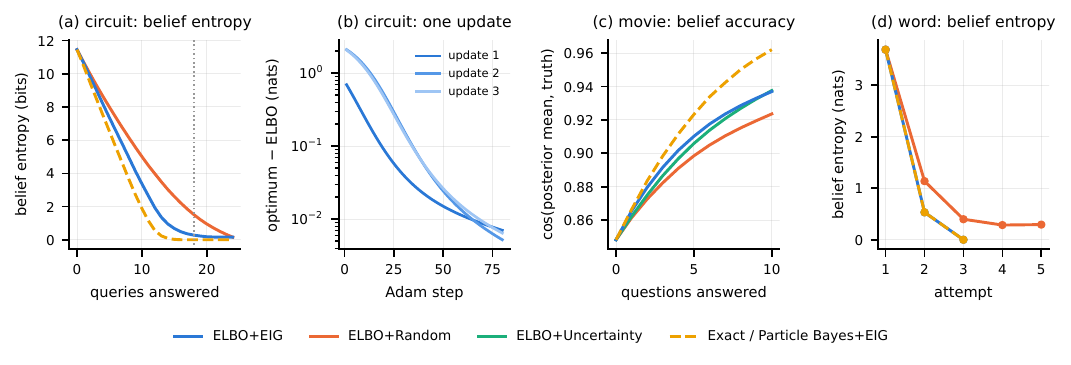}
\caption{Inference dynamics of the teachers. (a) Circuit Decoding: mean entropy of the belief after each query on the
synthetic training episodes (dotted line: the $18$-query budget). (b) One variational update on a training episode:
distance of the ELBO from its analytic optimum over the $80$ Adam steps, for the first three updates. (c) Movie Recommendation: cosine between the posterior mean and the true weights after each question (synthetic training
episodes). (d) Word Guess: mean belief entropy before each attempt on the complete test split (episodes still running;
the two EIG teachers coincide).}
\label{fig:teacher-dynamics}
\end{figure}

\textbf{Inference dynamics.} Figure~\ref{fig:teacher-dynamics} shows how the teachers' beliefs evolve. On Circuit Decoding the entropy of the variational belief falls from $11.5$ bits (the mean prior entropy over the $3{,}397$ training episodes, which are dominated by the two large inventories)
to $0.28$ bits after $18$ EIG queries, against $1.52$ bits with random queries and $0.00$ for exact inference with EIG
queries; the total-variation distance between the variational belief and the exact posterior is about $0.03$ at the
final turn ($0.027$ on the test split; $0.12$ under random queries, whose refuted hypotheses are eliminated more slowly). Within one update, the
$80$ Adam steps bring the ELBO to within about $0.01$ nats of its analytic optimum, the log marginal likelihood of the answer
under the previous belief, so the per-turn projection is essentially exact and the residual error is the smoothing
$\varepsilon$. On Movie Recommendation the cosine between the posterior mean and the true weights rises from $0.85$
(prior) to $0.94$ after ten EIG questions, $0.92$ with random questions and $0.96$ for the particle reference; the
remaining gap between the Gaussian and the particle belief is the approximation the student inherits. On Word Guess the
belief entropy drops from $3.69$ nats (uniform over $40$ words) to $0.53$ after one EIG attempt and to zero after two on
almost every episode, against $1.14$ and $0.40$ nats with random attempts; the variational update matches the analytic
categorical posterior to within a KL divergence of $6\times10^{-5}$ on every one of the $6{,}755$ updates.

\section{Synthetic Data and Distillation Details}
\label{app:data}

\subsection{Episode Generators and Separation from the Benchmark}
\label{app:generators}
\textbf{Circuit.} A layered generator over the public grammar with the three test inventories produces
$4{,}296$ episodes with distinct joint truth tables; each table is checked against the frozen test manifest
(overlap $0$). \textbf{Movie.} $220$ score tables $\times$ $20$ users $=4{,}400$ episodes; the generator
follows the format of the public rows (eight attributes, $20$ seen and $40$ unseen films, one- or two-decimal
scores sparse in three attributes) with attribute budgets calibrated to the public rows; users are drawn on
the same weight grid; the generator rejects the $1{,}000$ test weight vectors and the $20$ test score tables, and
every episode is checked against the frozen test manifest (weight-vector and score-table overlap $0$). Utilities are computed exactly (rounded to nine decimals) so that ties are handled as in
the benchmark. \textbf{Word.} $110$ vocabularies $\times$ $40$ secrets $=4{,}400$ episodes drawn from a pool
of $1{,}933$ five-letter words that excludes every word of the $10$ test vocabularies; every synthetic vocabulary
is checked to contain no test word.

\subsection{Data Splits and Dataset Statistics}
\label{app:splits}
\begin{center}\scriptsize
\setlength{\tabcolsep}{3pt}
\begin{tabular}{>{\raggedright\arraybackslash}p{4.6cm}ccc}
\toprule
 & Circuit & Movie & Word \\
\midrule
Episodes (train / val / held-out synthetic) & $3{,}397$ / $435$ / $464$ & $3{,}600$ / $400$ / $400$ & $3{,}600$ / $400$ / $400$ \\
Split unit & truth table (by inventory) & score table & vocabulary \\
Messages per dialogue & $40$ & $24$ & $2\times$attempts \\
Tokens per dialogue, action-only & $949$ (all) & $\approx3{,}570$ (max $3{,}624$) & $\le 971$ \\
Tokens per dialogue, belief scratchpad & $3{,}516$ (all) & max $4{,}709$ & max $1{,}388$ \\
Supervised tokens per teacher, action-only & $0.78$M & $0.82$M & $0.07$--$0.10$M \\
Supervised tokens per teacher, belief scratchpad & $8.85$M & -- & $2.83$M \\
Teacher success on training episodes (\ELBOEIG{}) & $1.000$ & $0.434$ & $2.31$ attempts \\
\bottomrule
\end{tabular}
\end{center}

\subsection{Dialogue Serialization for Action-Only Targets}
\label{app:serialisation}
One dialogue is one chat: the system prompt of Appendix~\ref{app:prompts}, a user message with the public
episode information (the inventory; the attribute names and film lists; the vocabulary), then alternating
assistant actions and user answers, the terminal user prompt, and the assistant's decision. Only assistant
messages are supervised. Circuit example (abridged):
\begin{verbatim}
user:      Circuit Decoding configuration: ... components: AND=3, OR=3, NOT=0.
           Ask your first query.
assistant: <query>C(0,0,1)</query>
user:      0
assistant: <query>C(1,0,1)</query>
...        (18 queries in total)
user:      Your query budget is exhausted. Submit your final answer now. ...
assistant: <guess>001101110001001100110111</guess>
\end{verbatim}
A validator checks every compiled dialogue for format, causality (no future action visible in any prefix),
absence of hidden-state leakage (the latent state, circuit wiring, utilities and diagnostics never appear) and
fidelity to the teacher trace.

\subsection{Belief-Scratchpad Formats}
\label{app:scratchpads}
\textbf{Movie.} Before every question: \texttt{<belief>w1, ..., w8</belief>} with the teacher's posterior
mean mapped to weight space, two decimals (the prior line is $0.45$ repeated eight times). Before the recommendation:
\texttt{<belief>...</belief>}, then \texttt{<scores>TITLE: S; ...</scores>} listing every unseen film in
the given order with its utility under the written weights (two decimals), then \texttt{<recommend>}. In the
training targets the recommendation is the argmax of the written scores in $99.3$ to $99.6\%$ of dialogues
(rounding accounts for the rest).

\textbf{Circuit.} Before every query and before the final guess: \texttt{<belief>b1, ..., b24</belief>} with the
teacher's posterior marginal that each of the $24$ truth-table bits equals one, two decimals, followed by
\texttt{<query>} or \texttt{<guess>}; the final instruction states that a guess bit is one exactly when its
belief entry exceeds $0.50$. The recorded teacher traces store the teacher's actions but not its beliefs, so the
marginals are recomputed by replaying each teacher's own inference on its recorded interactions and checked
against the recorded per-turn entropy, mass on the true table and number of consistent tables. The guess agrees
with the written final belief in every \ELBOEIG{} and \ExactEIG{} dialogue and in $62\%$ of the \ELBORand{}
dialogues: that teacher identifies the circuit in only $56\%$ of training episodes and submits a posterior
sample, which we leave unrepaired.

\textbf{Word.} From the second attempt on: \texttt{<feedback>S grey, P grey, I grey, T grey, E green</feedback>}
pairing each letter of the previous guess with its color; \texttt{<check>} listing every candidate of the
previous \texttt{<candidates>} line, spelled letter by letter, with \texttt{ok} or the first violated
constraint (\texttt{E green@5 -> out}); \texttt{<candidates>} listing the survivors spelled letter by letter;
then \texttt{<attempt>}. The first turn writes the full vocabulary as candidates. All lines are compiled
deterministically from the frozen teacher traces, so the scratchpad is exact by construction.

\subsection{Scratchpad Designs That Did Not Work}
\label{app:failed-scratchpads}
The first word-guess scratchpad listed the surviving candidates as whole words, without the feedback and
check lines. Its student imitated the first guess and the format perfectly but wrote a wrong candidate list
from the second turn on (equal to the exact set on $2$ of $98$ second turns; containing the secret in $16$),
then committed to the last surviving word and repeated it to the budget: $31.9$ attempts and $22\%$ solved on
$100$ test episodes, against $2.29$ for its teacher. The written list was a coarse function of the vocabulary
and the color counts, not of letter positions, which Llama's tokenizer never fully exposes (\texttt{ENACT} is
\texttt{EN|ACT}, \texttt{FROST} is \texttt{F|RO|ST}). The letter-spelled format of
Appendix~\ref{app:scratchpads} fixed this ($3.09$ attempts, $98\%$ solved on the same $100$ episodes). The
$100$ episodes are part of the $400$ reported in the paper; the format was fixed once, before the
remaining $300$ episodes were evaluated, and was run without further change; the complete test split and the
$400$ held-out synthetic episodes agree ($3.37$ against $3.37$ attempts).

\subsection{LLM-Teacher Baseline}
\label{app:llm-teacher}
Llama-3.1-70B-Instruct (bf16, tensor parallel $4$, vLLM~\citep{kwon2023vllm}) is rolled out greedily on the
same training and validation episodes as the Bayesian teachers, under the same prompts, with a token-trie
logits processor that admits only legal actions (unasked legal queries; a valid $24$-bit guess; legal pairs of
seen films; unseen titles), so its interactions are always well-formed. Its trace quality is measured with the
same reference posterior as everything else. Circuit ($3{,}397$ training episodes): Success@18 $0.002$,
bitwise accuracy $0.686$, EIG-optimal $0.300$, regret $0.340$ bits; it identifies the circuit by turn $18$ in
$45\%$ of episodes ($0.19$ GPU-hours on four A100s). Movie ($3{,}600$ episodes): Success@10 $0.078$,
normalized rank $0.409$, reference-EIG-optimal $0.002$, regret $0.269$ nats; it opens with the same pair in
$38\%$ of episodes. Its action-only dialogues are compiled and its students trained with the identical
recipe; their validation losses are the lowest of all students ($0.016$ circuit, $0.137$ movie), because a
greedy language model is the most predictable teacher, not the best one.

\subsection{Training Configuration and Compute}
\label{app:training}
\begin{center}\footnotesize
\begin{tabular}{lp{10.2cm}}
\toprule
Student & \Student{}, bf16, SDPA attention, gradient checkpointing \\
Adapters & LoRA $r=32$, $\alpha=64$, dropout $0.05$ on q, k, v, o, gate, up, down projections \\
Objective & next-token cross-entropy on assistant tokens (system, user and answer tokens masked) \\
Optimizer & AdamW~\citep{loshchilov2019adamw} (fused), learning rate $2\times10^{-4}$, cosine schedule, $3\%$ warm-up, weight decay $0.01$ \\
Batch & $2$ dialogues per GPU $\times$ $2$ GPUs $\times$ $4$ accumulation steps $=16$ dialogues per update \\
Epochs & $3$; checkpoint with the lowest validation loss on the synthetic validation split \\
Sequence length & $2{,}048$ (circuit action-only), $4{,}096$ (circuit scratchpad; movie action-only; word), $6{,}144$ (movie scratchpad); no truncation \\
Seed & $0$ for every condition \\
Wall time (2$\times$A100) & circuit action-only $32$ min, scratchpad $1.9$ h; movie action-only $2.6$ h, scratchpad $3.9$ h; word action-only $26$ min, scratchpad $47$ min \\
\bottomrule
\end{tabular}
\end{center}
With adapter parameters $\psi$ and a dialogue $x_{1:L}$ whose assistant-token positions form the set
$\mathcal T$, the objective is the next-token cross-entropy on those positions only,
\begin{equation}
\label{eq:sft}
\mathcal L_{\mathrm{SFT}}(\psi)=-\sum_{\text{dialogues}}\;\sum_{t\in\mathcal T}\log p_\psi(x_t\mid x_{<t}),
\end{equation}
averaged over the assistant tokens of a batch; system, user and answer tokens receive no loss.
Validation losses (epochs $1/2/3$): circuit \ELBOEIG{} $0.099/0.069/0.061$, \ELBORand{} $0.208/0.201/0.200$,
\ExactEIG{} $0.146/0.131/0.124$, \LLMteacher{} best $0.016$; circuit scratchpad \ELBOEIG{} $0.044/0.024/0.020$, \ELBORand{} $0.133/0.111/0.105$, \ExactEIG{} $0.017/0.012/0.012$; movie action-only \ELBOEIG{} $0.233/0.211/0.213$,
scratchpad \ELBOEIG{} $0.199/0.161/0.151$; word scratchpad \ELBOEIG{} $0.0144/0.0164/0.0167$ (best at epoch
$1$), word action-only \ELBOEIG{} $0.945/0.702/0.840$, \ELBORand{} $0.462/0.442/0.440$, \ExactEIG{}
$0.878/0.777/0.869$. Validation loss orders teachers by the predictability of their action sequences, not by their quality.

\begin{figure}[t]
\centering
\includegraphics[width=\linewidth]{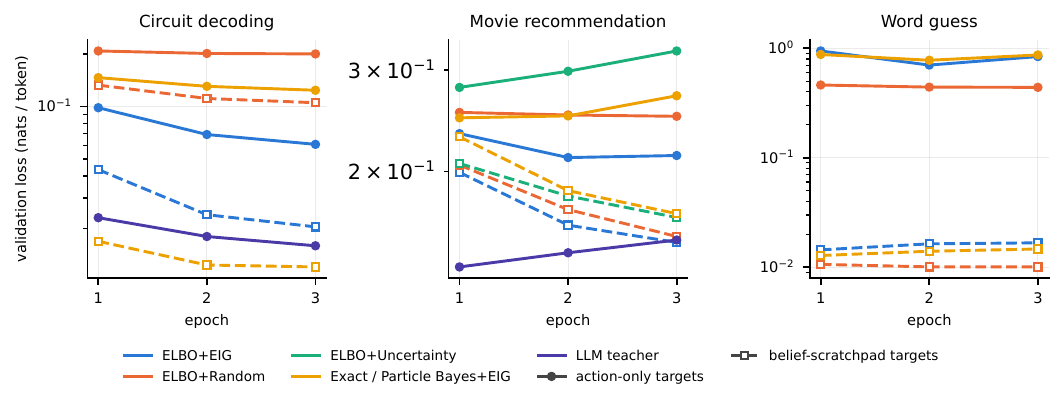}
\caption{Validation loss per epoch (synthetic validation split, assistant tokens) of every distilled student, one panel per
task. Solid lines with filled markers: action-only targets; dashed lines with open markers: belief-scratchpad targets.
Validation loss measures how predictable a teacher's interactions are, not how good they are: the \LLMteacher{} students
have the lowest loss and the worst benchmark scores, the action-only exact-inference circuit student has twice the loss of the corresponding
variational student, and the word action-only students' losses are more than an order of magnitude above the scratchpad
students', whose targets contain the deterministic bookkeeping. Losses are not comparable across target formats within a task
(different supervised tokens). The checkpoint with the lowest validation loss is the one evaluated.}
\label{fig:val-loss}
\end{figure}
Figure~\ref{fig:val-loss} shows the validation loss of every student per epoch.

\section{Evaluation Protocol and Statistical Methods}
\label{app:protocol}

\subsection{Protocol Specification and Integrity Checks}
Each evaluator is a single program per task that loads the base model and, for student conditions, one LoRA
adapter; runs the deterministic respondent; uses the frozen prompt template of the condition's target format,
greedy decoding (temperature $0$), and fixed per-turn token budgets (circuit $32$ per query and $64$ for the guess for action-only, $224$ and $256$
for scratchpad; movie $48$ and $32$ for action-only, $96$ and $640$ for scratchpad; word $256$ for action-only, $1{,}024$
for scratchpad). Before any episode runs, the program verifies SHA-256 hashes of the test manifest, the prompt
templates, the adapters and the frozen training data, and aborts on any mismatch; a four-episode smoke run
precedes the full run. Runs are resumable per episode and outputs are immutable once complete. Where a
condition's episodes had already been evaluated by an earlier run with byte-identical evaluator code, they
were reused after row-for-row verification and a fresh in-run reproduction of the smoke episodes.

\subsection{Handling of Invalid, Repeated and Malformed Outputs}
As in Appendix~\ref{app:rules}. Outputs are parsed by a strict parser: exactly one tag of the expected kind;
for scratchpad conditions the parser skips well-formed scratchpad blocks and parses the tagged action after
them, so a malformed scratchpad does not itself invalidate a turn.

\subsection{Definitions of the Reported Metrics}
\label{app:metric-defs}
Notation follows Appendix~\ref{app:notation}: an episode has true state $\theta^\ast$, query turns
$t=1,\dots,T$ with the model's action $\hat a_t$ and answer $y_t$, history $\hist_t$ and terminal decision $d$.
Unless stated otherwise, a rate is the mean of a per-turn or per-episode indicator over all episodes of a
condition.

\textbf{Task success and benchmark scores.} On Circuit Decoding $d\in\{0,1\}^{24}$ is the submitted table and
$d^{(c)}$, $\theta^{\ast(c)}$ are the eight-bit blocks of circuit $c\in\{A,B,C\}$; a malformed guess counts as
all bits wrong:
\begin{equation}
\label{eq:succ-circuit}
\begin{aligned}
\text{Success@18}&=\mathbf 1[d=\theta^\ast],\qquad
\text{bit acc.}=\frac{1}{24}\sum_{i=1}^{24}\mathbf 1[d_i=\theta^\ast_i],\\
\text{circuit-wise acc.}&=\frac{1}{3}\sum_{c\in\{A,B,C\}}\mathbf 1\big[d^{(c)}=\theta^{\ast(c)}\big],
\end{aligned}
\end{equation}
and the benchmark score is $100\times$ the mean circuit-wise accuracy. On Movie Recommendation, with
$u(i)=\theta^{\ast\top}\phi_i$ over the $40$ unseen films and $d$ the recommended film,
\begin{equation}
\label{eq:succ-movie}
\begin{aligned}
\mathrm{rank}(d)&=1+\big|\{i:\,u(i)>u(d)\}\big|,\qquad
\text{normalized rank}=\frac{\mathrm{rank}(d)-1}{39},\\
\text{Success@10}&=\mathbf 1\big[u(d)=\max\nolimits_i u(i)\big],
\end{aligned}
\end{equation}
so tied films share the best rank and any optimal film counts; the benchmark score is
$100\times(1-\text{mean normalized rank})$, and a malformed recommendation has normalized rank $1$ and
success $0$. On Word Guess, $N$ is the number of attempts up to and including the correct one, invalid and
repeated attempts included, with $N=40$ for an unsolved episode; mean attempts is the mean of $N$ and
$\text{Success@}k=\mathbf 1[N\le k]$.

\textbf{Reference posterior.} On the model's own history, and without ever showing it to the model, the evaluator
maintains
\begin{equation}
\label{eq:refpost}
b_t(\theta)\;\propto\;p(\theta)\prod_{s\le t}p(y_s\mid\theta,\hat a_s)
\end{equation}
over the turns with a legal action, with the task's deterministic likelihood and uniform prior: exactly, over
the $K$ grammar-enumerated tables (circuit) or the $40$ words (word), where $b_t$ is uniform on the
\emph{version space} $V_t=\{\theta:\,p(y_s\mid\theta,\hat a_s)=1\text{ for all }s\le t\}$ (the consistent set
$C_t$ on Word Guess); and on Movie Recommendation by the particle posterior of
Appendix~\ref{app:teacher-refs}, $N_p=200{,}000$ particles $\theta^{(n)}$ drawn uniformly from the grid
$\{0,0.1,\dots,0.9\}^8$ with seed $0$ and weights
\begin{equation}
\label{eq:refpost-movie}
w_t^{(n)}\;\propto\;\prod_{s\le t}\sigma\!\Big(z_s\,\frac{u^{(n)}(i_s)-u^{(n)}(j_s)}{\tau}\Big),\qquad \tau=0.1,
\end{equation}
where $u^{(n)}(i)=\theta^{(n)\top}\phi_i$, $(i_s,j_s)$ is the pair asked at turn $s$ and $z_s=+1$ for Yes and
$-1$ for No; No Preference answers and invalid turns contribute no factor, and the answer to a repeated pair is
applied once more.

\textbf{Reference EIG, EIG-optimal rate and EIG regret.} At turn $t$ the admissible set $\sA_t$ of
Eq.~\ref{eq:eig} is the set of unasked queries (circuit), unguessed words (word) or all $190$ pairs (movie).
The reference EIG of $a\in\sA_t$ is Eq.~\ref{eq:eig} evaluated under $b_{t-1}$ instead of $q_{t-1}$,
\begin{equation}
\label{eq:refeig}
\EIG^{\mathrm{ref}}_t(a)=\Hent\big[b_{t-1}(y\mid a)\big]-\E_{b_{t-1}(\theta)}\Hent\big[p(y\mid\theta,a)\big],
\qquad b_{t-1}(y\mid a)=\E_{b_{t-1}(\theta)}\,p(y\mid\theta,a),
\end{equation}
which for a deterministic likelihood is the entropy of the partition of $b_{t-1}$ induced by the possible
answers: the binary entropy of the predicted bit, in bits, on Circuit Decoding, and the entropy over the
feedback patterns, in nats, on Word Guess; on Movie Recommendation it is evaluated under the particle
weights, in nats. With $\EIG^\ast_t=\max_{a\in\sA_t}\EIG^{\mathrm{ref}}_t(a)$,
\begin{equation}
\label{eq:eigopt}
o_t=\mathbf 1\big[\EIG^{\mathrm{ref}}_t(\hat a_t)\ge\EIG^\ast_t-\epsilon\big],\qquad
r_t=\EIG^\ast_t-\EIG^{\mathrm{ref}}_t(\hat a_t),
\end{equation}
with $\epsilon=10^{-9}$ bits (circuit), $10^{-9}$ nats (word) and $10^{-6}$ nats (movie), so that ties count
as optimal. The \emph{EIG-optimal rate} is the mean of $o_t$ and the \emph{EIG regret} the mean of $r_t$ over
the counted query turns of all episodes. Wasted turns: on Circuit Decoding an invalid or repeated query is
scored with $\EIG^{\mathrm{ref}}_t(\hat a_t)=0$, so $o_t=0$ and $r_t=\EIG^\ast_t$; on Movie Recommendation an
invalid question is scored the same way and a repeated pair like any legal pair under $b_{t-1}$; on Word Guess
invalid and repeated attempts are excluded from both means (they enter $N$).

\textbf{Identification and posterior mass.} The posterior mass on the true state is $b_t(\theta^\ast)$. A
history identifies the state when the version space is a singleton,
\begin{equation}
\label{eq:ident}
\begin{aligned}
\text{identified@}T&=\mathbf 1\big[|V_T|=1\big]\ \ (\text{circuit, word}),\\
\text{turns to identify}&=\min\{t:\,\theta^{\mathrm{MAP}}_s=\theta^\ast\ \text{for all}\ s\ge t\},
\end{aligned}
\end{equation}
where the second quantity, reported for the circuit teachers, is the first turn after which the teacher's
maximum-a-posteriori table $\theta^{\mathrm{MAP}}_s$ stays correct through the last of its $24$ diagnostic
queries ($25$ if it never does). On Movie Recommendation the analogous quantities use the mass on the true
best film,
\begin{equation}
\label{eq:pbest}
p_{\mathrm{best},t}=\sum_{n}w_t^{(n)}\,\mathbf 1\Big[\arg\max\nolimits_i u^{(n)}(i)\in B^\ast\Big],\qquad
\text{identified@10}=\mathbf 1\big[p_{\mathrm{best},10}\ge0.9\big],
\end{equation}
with $B^\ast$ the set of unseen films attaining $\max_i u(i)$. The \emph{oracle read-out} of
Appendix~\ref{app:extra} is the posterior mode of the best film under $b_{10}$ (movie) and, on Word Guess,
the indicator that $|C_t|=1$ for some $t\le k$.

\textbf{Interaction quality.} The \emph{invalid rate} is the fraction of query turns whose output fails the
strict parser or names an illegal action, and the \emph{repeat rate} the fraction whose output is a legal
action already taken in the episode.

\textbf{Belief and read-out fidelity} (scratchpad students). A scratchpad is \emph{well formed} when its
belief block parses under the strict parser. On Circuit Decoding the written marginals are
$\hat b_t\in[0,1]^{24}$ and
\begin{equation}
\label{eq:fid-circuit}
\begin{aligned}
\mathrm{MAE}_t&=\frac{1}{24}\sum_{i=1}^{24}\big|\hat b_{t,i}-b_t(\theta_i{=}1)\big|,\\
\text{guess consistent}&=\mathbf 1\Big[d_i=\mathbf 1[\hat b_{T,i}>0.5]\ \text{for all}\ i\ \text{with}\
|\hat b_{T,i}-0.5|>0.005\Big],
\end{aligned}
\end{equation}
where $b_t(\theta_i{=}1)$ is the exact marginal of bit $i$ under the reference posterior given the student's
own queries. On Movie Recommendation, with written weights $\hat\theta_t$ and written scores $\hat s(i)$ of
the unseen films: recommendation consistent $=\mathbf 1[d=\arg\max_i\hat s(i)]$; written belief optimal
$=\mathbf 1[\arg\max_i\hat\theta_T^{\top}\phi_i\in B^\ast]$; distance to the reference
$=\|\hat\theta_T-\E_{b_T}[\theta]\|_2$ and to the truth $=\|\hat\theta_T-\theta^\ast\|_2$. On Word Guess, for
turns $t\ge2$ with written candidate list $\hat C_t$: candidates exact $=\mathbf 1[\hat C_t=C_t]$; secret
retained $=\mathbf 1[\theta^\ast\in\hat C_t]$; feedback line correct $=1$ when the written (letter, color)
pairs equal the true feedback of the previous attempt; check verdicts correct $=$ the fraction of listed
candidates whose written verdict (kept or eliminated) agrees with the two-pass feedback rule applied to the
previous attempt.

\subsection{Statistical Tests and Confidence Intervals}
\label{app:stats}
All comparisons are paired on identical episodes. Binary metrics: exact binomial McNemar test on discordant pairs; paired
bootstrap of the mean difference with $10{,}000$ resamples of episodes (seed $0$), reporting the $2.5$ and
$97.5$ percentiles. Attempts: two-sided Wilcoxon signed-rank test on non-zero differences and the same paired
bootstrap; a cluster bootstrap resamples the $10$ vocabularies. Per-condition CIs of a proportion are Wilson
intervals or the same bootstrap.

\subsection{Validation Checks Prior to Benchmark Evaluation}
\label{app:gates}
The checks against rules fixed in advance that preceded a benchmark run were run on $200$ held-out synthetic
episodes for the belief-scratchpad students of Movie Recommendation and Circuit Decoding and on the first $100$
test episodes of Appendix~\ref{app:failed-scratchpads} for the \ELBOEIG{} letter-scratchpad student of Word
Guess; the circuit action-only students had teacher-level and smoke checks only. No other student was checked
against a pre-specified rule before its benchmark run; every run had the integrity and smoke checks of
Appendix~\ref{app:protocol}. Movie ($200$ held-out episodes, rule: paired gain of the scratchpad student over its action-only
counterpart $\ge+0.10$ and recommendation equal to the argmax of the written scores in $\ge80\%$ of episodes):
\ELBOEIG{} $0.320$ against $0.105$ ($+0.215$ \ci{+0.140}{+0.290}), \ELBORand{} $0.230$ against $0.130$,
\ELBOUnc{} $0.235$ against $0.080$, \ParticleEIG{} $0.415$ against $0.110$; the subsequent benchmark values
$0.345$ / $0.176$ / $0.194$ / $0.409$ preserve the ordering, and each lies inside or next to the held-out
interval. Word (rule applied on the first $100$ test episodes, the same $100$ on which the scratchpad format was fixed,
Appendix~\ref{app:failed-scratchpads}; rule: mean attempts $\le6$ and exact candidate tracking on turns $\ge2$
in $\ge80\%$): the letter-scratchpad student passed the first criterion ($3.09$ attempts, $98$ of $100$
solved) and the second on non-repeated turns ($0.81$; $0.53$ when the repeated turns of the two unsolved
episodes are pooled in), which we report as written. Circuit,
action-only students: the teacher-ordering check on the synthetic episodes (\ELBOEIG{} $\approx$ \ExactEIG{}
$\gg$ \ELBORand{}) and the training smoke checks were the only pre-benchmark rules. Circuit, belief-scratchpad
students ($200$ held-out episodes, all seven conditions on the same episodes; rule for the \ELBOEIG{} pair,
fixed before any of these students was trained: a well-formed belief on $\ge95\%$ of query turns, a parsable guess
in $\ge95\%$ of episodes, and Success@18 no more than $0.10$ below the action-only student): \ELBOEIG{} $0.910$
against $0.920$ for its action-only counterpart ($-0.010$ \ci{-0.055}{+0.035}), \ELBORand{} $0.530$ against
$0.485$ ($+0.045$ \ci{-0.055}{+0.150}), \ExactEIG{} $0.735$ against $0.625$ ($+0.110$ \ci{+0.015}{+0.205},
McNemar $p=0.030$); beliefs were well formed on at least $99.9\%$ of query turns and the written final marginals
lie within a mean absolute error of $0.007$, $0.034$ and $0.015$ per bit of the exact posterior given the
student's own queries. The action-only students approximately reproduce their benchmark values on these episodes ($0.920$,
$0.485$, $0.625$ against $0.920$, $0.500$, $0.643$). Word students were additionally evaluated on all $400$
held-out synthetic episodes after the benchmark run (Appendix~\ref{app:word-extra}).

\subsection{Base-Model Behavior under Each Prompt Format}
\label{app:base-behaviour}
Circuit: $493$ unparsable queries, $97$ of $300$ malformed final guesses, bitwise accuracy $0.380$, EIG-optimal
$0.239$; the base's queries happen to identify the circuit in $33\%$ of episodes and it never reads the table
out. Movie, action-only prompt: every question is invalid (untagged or malformed); the recommendation parses
in $100\%$ of episodes and is near chance ($0.032$; normalized rank $0.496$). Movie, scratchpad prompt: the
base copies the prior belief line, writes an untagged question and an all-zero scores line without a tagged
recommendation, and scores $0$ by construction. Word: $17\%$ invalid attempts (six-letter words, non-words),
$58\%$ repeats, $32.0$ attempts, yet its guesses isolate the secret within three attempts under the exact
consistency check in $80\%$ of episodes: the base fails more at reading the feedback out than at choosing informative words.

\section{Additional Distillation Results}
\label{app:extra}
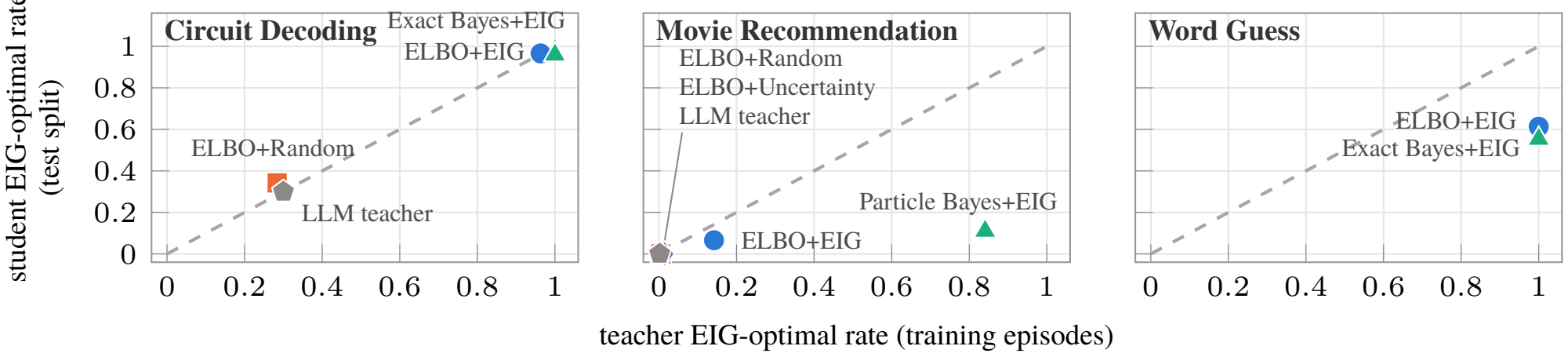
\begin{figure}[htbp]
\centering
\definecolor{vizblue}{HTML}{2a78d6}
\definecolor{vizorange}{HTML}{eb6834}
\definecolor{vizaqua}{HTML}{1baf7a}
\definecolor{vizpurple}{HTML}{8a5cd6}
\definecolor{vizgray}{HTML}{8a8985}
\pgfplotsset{
  diag/.style={black!35, dashed, thick, forget plot},
  teig/.style={only marks, mark=*, mark size=2.6pt, color=vizblue},
  trand/.style={only marks, mark=square*, mark size=2.5pt, color=vizorange},
  tunc/.style={only marks, mark=diamond*, mark size=3.1pt, color=vizpurple},
  tbayes/.style={only marks, mark=triangle*, mark size=3.1pt, color=vizaqua},
  tllm/.style={only marks, mark=pentagon*, mark size=2.9pt, color=vizgray},
}
\tikzset{lab/.style={font=\tiny, text=black!70, inner sep=1pt},
         leader/.style={black!45, line width=0.4pt},
         panel/.style={font=\scriptsize\bfseries, text=black!80, anchor=north west, inner sep=1pt}}
\begin{tikzpicture}
\begin{groupplot}[
  group style={group size=3 by 1, horizontal sep=0.55cm, yticklabels at=edge left},
  scale only axis, width=3.6cm, height=2.1cm,
  xmin=-0.04, xmax=1.06, ymin=-0.04, ymax=1.16,
  xtick={0,0.2,0.4,0.6,0.8,1.0}, ytick={0,0.2,0.4,0.6,0.8,1.0},
  xtick pos=bottom, ytick pos=left,
  tick label style={font=\scriptsize}, label style={font=\scriptsize},
  grid=major, grid style={black!10}, axis line style={black!40},
  every axis plot/.append style={mark options={draw=white, line width=0.4pt}},
  clip=false,
]
\nextgroupplot[ylabel={student EIG-optimal rate\\(test split)}, ylabel style={align=center}]
\node[panel] at (axis cs:-0.02,1.15) {Circuit Decoding};
\addplot[diag] coordinates {(0,0) (1,1)};
\addplot[teig] coordinates {(0.963,0.966)};
\addplot[trand] coordinates {(0.284,0.342)};
\addplot[tbayes] coordinates {(1.000,0.959)};
\addplot[tllm] coordinates {(0.300,0.300)};
\node[lab, anchor=east] at (axis cs:0.935,0.975) {ELBO+EIG};
\draw[leader] (axis cs:1.0,0.99) -- (axis cs:1.0,1.03);
\node[lab, anchor=south east] at (axis cs:1.04,1.035) {Exact Bayes+EIG};
\node[lab, anchor=south west] at (axis cs:0.05,0.445) {ELBO+Random};
\node[lab, anchor=north west] at (axis cs:0.335,0.265) {LLM teacher};
\nextgroupplot[xlabel={teacher EIG-optimal rate (training episodes)}]
\node[panel] at (axis cs:-0.02,1.15) {Movie Recommendation};
\addplot[diag] coordinates {(0,0) (1,1)};
\addplot[teig] coordinates {(0.142,0.066)};
\addplot[trand] coordinates {(0.005,0.000)};
\addplot[tunc] coordinates {(0.015,0.000)};
\addplot[tbayes] coordinates {(0.841,0.110)};
\addplot[tllm] coordinates {(0.002,0.001)};
\node[lab, anchor=west] at (axis cs:0.20,0.066) {ELBO+EIG};
\node[lab, anchor=south east] at (axis cs:1.04,0.165) {Particle Bayes+EIG};
\draw[leader] (axis cs:0.012,0.03) -- (axis cs:0.06,0.585);
\node[lab, anchor=south west, align=left] at (axis cs:0.04,0.60) {ELBO+Random\\ELBO+Uncertainty\\LLM teacher};
\nextgroupplot
\node[panel] at (axis cs:-0.02,1.15) {Word Guess};
\addplot[diag] coordinates {(0,0) (1,1)};
\addplot[teig] coordinates {(1.000,0.613)};
\addplot[tbayes] coordinates {(1.000,0.553)};
\node[lab, anchor=east] at (axis cs:0.955,0.645) {ELBO+EIG};
\node[lab, anchor=east] at (axis cs:0.965,0.50) {Exact Bayes+EIG};
\end{groupplot}
\end{tikzpicture}
\caption{Students inherit their teacher's information-seeking policy on Circuit Decoding and only partly on
Movie Recommendation and Word Guess: the teacher's EIG-optimal rate on the training episodes against its
student's on the complete test split, one panel per task, one labeled point per teacher (belief-scratchpad
students; the \LLMteacher{}'s students are action-only).}
\label{fig:inherit}
\end{figure}

\subsection{Circuit Decoding}
\label{app:circuit-extra}
\begin{center}\scriptsize
\setlength{\tabcolsep}{2pt}
\begin{tabular}{lcccccccc}
\toprule
Condition & Success@18 & bit acc. & final parse & valid query & EIG-opt. & regret & identified@18 & val.\ loss \\
\midrule
Base & $0.000$ & $0.380$ & $0.677$ & $0.909$ & $0.239$ & $0.355$ & $0.330$ & -- \\
\ELBORand{} student & $0.500$ & $0.955$ & $1.000$ & $1.000$ & $0.310$ & $0.310$ & $0.567$ & $0.200$ \\
\ELBOEIG{} student & $\mathbf{0.920}$ & $\mathbf{0.996}$ & $1.000$ & $1.000$ & $\mathbf{0.954}$ & $\mathbf{0.014}$ & $\mathbf{0.997}$ & $0.061$ \\
\ExactEIG{} student & $0.643$ & $0.974$ & $1.000$ & $1.000$ & $0.896$ & $0.066$ & $0.897$ & $0.124$ \\
\LLMteacher{} student & $0.000$ & $0.619$ & $1.000$ & $1.000$ & $0.300$ & $0.340$ & -- & $0.016$ \\
\midrule
\ELBORand{} student (B) & $0.590$ & $0.972$ & $1.000$ & $1.000$ & $0.342$ & $0.291$ & $0.713$ & $0.105$ \\
\ELBOEIG{} student (B) & $\mathbf{0.927}$ & $\mathbf{0.996}$ & $1.000$ & $0.996$ & $\mathbf{0.966}$ & $\mathbf{0.008}$ & $\mathbf{0.993}$ & $0.020$ \\
\ExactEIG{} student (B) & $0.840$ & $0.989$ & $1.000$ & $1.000$ & $0.959$ & $0.027$ & $0.963$ & $0.012$ \\
\bottomrule
\end{tabular}
\end{center}
Rows marked (B) are the belief-scratchpad students and all other rows are action-only students; validation losses are
not comparable across the two formats because the scratchpad tokens are supervised. The paragraphs below
describe the action-only students unless stated otherwise.
Per inventory ($3/3/0$, $3/3/1$, $1/5/2$), Success@18: \ELBOEIG{} $1.00$ / $0.84$ / $0.92$; \ExactEIG{} $0.85$
/ $0.52$ / $0.56$; \ELBORand{} $0.56$ / $0.42$ / $0.52$; \LLMteacher{} student $0$ / $0$ / $0$.
\textbf{Final-answer errors.} \ELBOEIG{} student: $24$ failures, $23$ after the version space had collapsed to
one table (off by one bit in $17$, two in $5$, three in $1$) and $1$ with two hypotheses left; regret $0.041$ in failures against $0.012$ in
successes. \ExactEIG{} student: $107$ failures, $89$ after collapse ($36$ off by one bit, $45$ by two, $8$ by
three or four), $18$ with two hypotheses left; queries are exactly optimal on turns $1$--$5$ and $0.80$--$0.95$
optimal afterward. \ELBORand{} student: $150$ failures, $93$ with at least two consistent hypotheses left at
the final guess ($77$ with exactly two) and $57$ read-out errors after collapse. Opening queries are
deterministic per inventory under greedy decoding (the prompt differs only by inventory before the first
query). \textbf{Complete paired table} (same $300$ episodes; first-named minus second-named condition; discordant counts first-only / second-only):
\ELBOEIG{} vs.\ \ELBORand{} $+0.420$ \ci{+0.350}{+0.487}, $143/17$, $p=5.3\times10^{-26}$; \ExactEIG{} vs.\
\ELBOEIG{} $-0.277$ \ci{-0.333}{-0.217}, $9/92$, $p=1.8\times10^{-18}$; \ExactEIG{} vs.\ \ELBORand{} $+0.143$
\ci{+0.067}{+0.220}, $94/51$, $p=4.5\times10^{-4}$; \ELBOEIG{} vs.\ base $+0.920$ \ci{+0.887}{+0.950}, $276/0$;
every student vs.\ base: no discordant pair in the base's favor; \LLMteacher{} student vs.\ base $0/0$, $p=1$.
\textbf{Inheritance of the query policy}: EIG-optimal rate of the teacher on the
training episodes against its action-only student's on the test split, $0.963$ / $0.954$ (\ELBOEIG{}), $0.284$ / $0.310$
(\ELBORand{}), $1.000$ / $0.896$ (\ExactEIG{}) and $0.300$ / $0.300$ (\LLMteacher{}).

\textbf{Belief-scratchpad students} (same $300$ episodes, evaluated in a separate run and paired by episode).
Per inventory, Success@18: \ELBOEIG{} $1.00$ / $0.89$ / $0.89$; \ExactEIG{} $1.00$ / $0.70$ / $0.82$;
\ELBORand{} $0.79$ / $0.46$ / $0.52$. Format contrasts, scratchpad minus action-only (discordant counts,
solved only by the scratchpad student / only by the action-only student): \ELBOEIG{} $+0.007$ \ci{-0.037}{+0.050}, $22/20$, $p=0.88$; \ELBORand{}
$+0.090$ \ci{+0.010}{+0.170}, $89/62$, $p=0.034$; \ExactEIG{} $+0.197$ \ci{+0.133}{+0.257}, $82/23$,
$p=6.0\times10^{-9}$. Among the scratchpad students: \ELBOEIG{} vs.\ \ELBORand{} $+0.337$ \ci{+0.277}{+0.393},
$109/8$, $p=8.9\times10^{-24}$; \ExactEIG{} vs.\ \ELBOEIG{} $-0.087$ \ci{-0.137}{-0.037}, $16/42$,
$p=8.6\times10^{-4}$; \ExactEIG{} vs.\ \ELBORand{} $+0.250$ \ci{+0.187}{+0.313}, $95/20$, $p=7.1\times10^{-13}$;
every student vs.\ base: no discordant pair in the base's favor. The written belief is well formed on at least
$99.8\%$ of query turns and $99.7\%$ of final turns, the written final marginals lie within a mean absolute
error of $0.005$, $0.028$ and $0.011$ per bit of the exact posterior given the student's own queries
(\ELBOEIG{}, \ELBORand{}, \ExactEIG{}), and the guess equals the rounded written belief in $97\%$, $76\%$ and
$100\%$ of episodes. The final-answer errors keep their character. \ELBOEIG{} student: $22$ failures, $20$ after
the version space had collapsed ($14$ off by one bit). \ExactEIG{} student: $48$ failures against $107$ with
action-only targets, $40$ after collapse ($27$ off by one bit) and $8$ with two hypotheses left; its queries
are exactly optimal on turns $1$--$5$ and $0.89$ optimal afterward. \ELBORand{} student: $123$ failures, $64$
with at least two consistent hypotheses left ($53$ with exactly two) and $59$ read-out errors after collapse.
The scratchpad makes the read-out error visible: in all $40$ after-collapse failures of the \ExactEIG{} student
and in $15$ of the $20$ for the \ELBOEIG{} student, the erroneous bit is already wrong in the written final
belief, which the guess then copies ($40$ and $12$ episodes). The only invalid or repeated queries of any
circuit student are $22$ of the $5{,}400$ turns of the \ELBOEIG{} scratchpad student, all at turn $14$ or later
and after the circuit had been identified. Inheritance of the query policy (the circuit points of
Figure~\ref{fig:inherit}), teacher on the training episodes / scratchpad student on the test split: $0.963$ /
$0.966$ (\ELBOEIG{}), $0.284$ / $0.342$ (\ELBORand{}), $1.000$ / $0.959$ (\ExactEIG{}).

\subsection{Movie Recommendation}
\label{app:movie-extra}
\textbf{All students on the test split.}
\begin{center}\scriptsize
\setlength{\tabcolsep}{2.5pt}
\begin{tabular}{llcccccccc}
\toprule
Target & Teacher & Success@10 [95\% CI] & rank & bench.\ \% & repeat & pairs & EIG-opt. & regret & $p$(best) \\
\midrule
-- & Base & $0.032$ [$0.022$, $0.043$] & $0.496$ & $50.4$ & $0.000$ & $0.00$ & $0.000$ & $0.554$ & $0.075$ \\
A & \ELBOEIG{} & $0.085$ [$0.068$, $0.103$] & $0.396$ & $60.4$ & $0.509$ & $4.91$ & $0.025$ & $0.279$ & $0.172$ \\
A & \ELBORand{} & $0.105$ [$0.086$, $0.124$] & $0.390$ & $61.0$ & $0.801$ & $1.99$ & $0.000$ & $0.408$ & $0.112$ \\
A & \ELBOUnc{} & $0.064$ [$0.049$, $0.080$] & $0.423$ & $57.7$ & $0.583$ & $4.17$ & $0.000$ & $0.331$ & $0.156$ \\
A & \ParticleEIG{} & $0.079$ [$0.063$, $0.096$] & $0.398$ & $60.2$ & $0.040$ & $9.43$ & $0.024$ & $0.180$ & $0.263$ \\
A & \LLMteacher{} & $0.052$ [$0.039$, $0.066$] & $0.433$ & $56.7$ & $0.005$ & $9.91$ & $0.001$ & $0.276$ & $0.203$ \\
B & \ELBOEIG{} & $0.345$ [$0.316$, $0.375$] & $0.100$ & $90.0$ & $0.301$ & $6.99$ & $0.066$ & $0.162$ & $0.247$ \\
B & \ELBORand{} & $0.176$ [$0.153$, $0.200$] & $0.208$ & $79.2$ & $0.798$ & $2.02$ & $0.000$ & $0.399$ & $0.118$ \\
B & \ELBOUnc{} & $0.194$ [$0.170$, $0.219$] & $0.208$ & $79.2$ & $0.587$ & $4.13$ & $0.000$ & $0.332$ & $0.159$ \\
B & \ParticleEIG{} & $\mathbf{0.409}$ [$0.379$, $0.439$] & $\mathbf{0.069}$ & $\mathbf{93.1}$ & $0.044$ & $9.49$ & $0.110$ & $0.078$ & $0.317$ \\
\bottomrule
\end{tabular}
\end{center}
\textbf{Action-only contrasts} (paired, $1{,}000$ episodes): \ELBOEIG{} vs.\ \ELBORand{} $-0.020$
\ci{-0.038}{-0.003} ($p=0.035$); \ELBOEIG{} vs.\ \ELBOUnc{} $+0.021$ ($p=0.028$); \ParticleEIG{} vs.\ \ELBOEIG{}
$-0.006$ ($p=0.65$); \LLMteacher{} vs.\ \ELBOEIG{} $-0.033$ \ci{-0.055}{-0.011} ($p=4.7\times10^{-3}$).
\textbf{Scratchpad against action-only on identical episodes}: \ELBOEIG{} $+0.260$ \ci{+0.228}{+0.292}
($300/40$ discordant, $p=2.1\times10^{-50}$); \ELBORand{} $+0.071$ \ci{+0.044}{+0.098}; \ELBOUnc{} $+0.130$
\ci{+0.103}{+0.157}; \ParticleEIG{} $+0.330$ \ci{+0.296}{+0.362} ($365/35$, $p=2.1\times10^{-70}$); against
the action-only base: $+0.313$, $+0.144$, $+0.162$, $+0.377$ \ci{+0.345}{+0.409}.

\textbf{Read-out decomposition} (Section~\ref{sec:res-belief}). A student is given a teacher's ten questions
and answers and produces only the recommendation; the oracle read-out is the posterior mode of the best unseen
film under the reference posterior on the same transcript. Own questions + own read-out / own + oracle /
teacher questions + own read-out / teacher + oracle / teacher + teacher read-out: \ELBOEIG{} $0.085$ / $0.305$ /
$0.085$ / $0.473$ / $0.410$; \ELBORand{} $0.105$ / $0.195$ / $0.108$ / $0.413$ / $0.373$; \ELBOUnc{} $0.064$ /
$0.262$ / $0.060$ / $0.485$ / $0.428$; \ParticleEIG{} $0.079$ / $0.448$ / $0.077$ / $0.571$ / $0.562$;
\LLMteacher{} student $0.052$ / $0.330$ / $0.048$ / $0.571$ / $0.562$ (particle prefix). Under an oracle
read-out the teacher ordering re-emerges. \textbf{Attribute-permutation probe} ($200$ test episodes): the
attribute rows of the $40$ unseen films are permuted among the unchanged titles; a utility-reading model
follows the attributes to their new title, whereas a title-reading model keeps the title. Action-only students keep
the title in $7$--$25\%$ of episodes (base $38$--$42\%$), follow the attributes in $4$--$22\%$, and move to a
third title in $66$--$81\%$; success stays at chance under the permutation. \textbf{Belief fidelity}
(scratchpad students): belief and scores lines parse on $\ge98\%$ of turns; recommendation $=$ argmax of
written scores in $0.976$ / $0.999$ / $0.930$ / $0.966$ of episodes; argmax of written belief is optimal in
$0.338$ / $0.177$ / $0.201$ / $0.414$ (Success@10 $0.345$ / $0.176$ / $0.194$ / $0.409$); Euclidean distance
of the written final belief to the reference posterior mean $0.210$ / $0.087$ / $0.351$ / $0.135$, to the true
weights $0.633$ / $0.748$ / $0.785$ / $0.532$ (reference mean to true weights $0.591$ / $0.739$ / $0.683$ /
$0.506$). \textbf{Repeated questions}: no movie teacher excludes asked pairs (teacher repeat rates $0.014$--$0.064$,
Appendix~\ref{app:teacher-extended}), yet the \ELBORand{} and \ELBOUnc{} students repeat $58$--$80\%$ of their
questions under both formats, the \ELBOEIG{} student $30$--$51\%$ and the particle student $4\%$.

\subsection{Word Guess}
\label{app:word-extra}
\textbf{Test split} ($400$ episodes) and \textbf{held-out synthetic episodes} ($400$).
\begin{center}\scriptsize
\setlength{\tabcolsep}{3pt}
\begin{tabular}{llcccccccc}
\toprule
Episodes & Condition & attempts [95\% CI] & median & S@2 & S@3 & S@5 & unsolved & invalid & repeat \\
\midrule
test & Base & $31.97$ [$30.56$, $33.36$] & $40$ & $0.043$ & $0.060$ & $0.120$ & $300$ & $0.172$ & $0.577$ \\
test & \ELBOEIG{} A & $19.16$ [$17.52$, $20.84$] & $9$ & $0.077$ & $0.142$ & $0.318$ & $155$ & $0.047$ & $0.560$ \\
test & \ELBORand{} A & $15.23$ [$13.94$, $16.56$] & $9$ & $0.075$ & $0.130$ & $0.280$ & $79$ & $0.018$ & $0.243$ \\
test & \ExactEIG{} A & $15.37$ [$13.97$, $16.83$] & $8$ & $0.085$ & $0.155$ & $0.312$ & $98$ & $0.014$ & $0.432$ \\
test & \ELBOEIG{} B & $3.37$ [$2.87$, $3.95$] & $2$ & $0.505$ & $\mathbf{0.917}$ & $0.975$ & $9$ & $0.028$ & $0.221$ \\
test & \ELBORand{} B & $\mathbf{3.15}$ [$3.03$, $3.27$] & $3$ & $0.295$ & $0.713$ & $0.963$ & $\mathbf{0}$ & $0.000$ & $0.005$ \\
test & \ExactEIG{} B & $\mathbf{3.10}$ [$2.69$, $3.58$] & $2$ & $0.507$ & $\mathbf{0.917}$ & $0.983$ & $6$ & $0.045$ & $0.137$ \\
\midrule
synthetic & \ELBOEIG{} A & $17.69$ [$16.13$, $19.28$] & $9$ & $0.083$ & $0.158$ & $0.273$ & $134$ & $0.007$ & $0.534$ \\
synthetic & \ELBORand{} A & $14.92$ [$13.67$, $16.15$] & $10$ & $0.068$ & $0.125$ & $0.245$ & $71$ & $0.006$ & $0.188$ \\
synthetic & \ExactEIG{} A & $15.89$ [$14.45$, $17.37$] & $9$ & $0.087$ & $0.155$ & $0.318$ & $105$ & $0.010$ & $0.427$ \\
synthetic & \ELBOEIG{} B & $3.37$ [$2.86$, $3.97$] & $2$ & $0.522$ & $0.917$ & $0.970$ & $9$ & $0.013$ & $0.232$ \\
synthetic & \ELBORand{} B & $3.78$ [$3.27$, $4.36$] & $3$ & $0.357$ & $0.750$ & $0.958$ & $9$ & $0.000$ & $0.237$ \\
synthetic & \ExactEIG{} B & $3.00$ [$2.62$, $3.45$] & $2$ & $0.527$ & $0.915$ & $0.988$ & $5$ & $0.033$ & $0.129$ \\
\bottomrule
\end{tabular}
\end{center}
Teachers on the same episodes: $2.29$ / $3.09$ / $2.29$ attempts (test), $2.31$ / $3.20$ / $2.31$ (synthetic).
On solved episodes alone the EIG students take $2.53$ and $2.54$ attempts against $3.15$ for \ELBORand{}; the
difference in the means is entirely the $40$-attempt dead ends. \textbf{Paired contrasts} (test split):
\ELBOEIG{} B vs.\ base $-28.60$ \ci{-30.09}{-27.14}, Wilcoxon $p=2.1\times10^{-65}$; B vs.\ A for \ELBOEIG{} $-15.78$
\ci{-17.54}{-14.04}, Success@3 $313/3$ discordant, for \ELBORand{} $-12.08$ \ci{-13.42}{-10.78}, $239/6$, and for
\ExactEIG{} $-12.27$ \ci{-13.80}{-10.82}, $311/6$ (held-out synthetic episodes: $-14.32$, $-11.14$, $-12.90$); \ELBOEIG{} B vs.\ \ELBORand{} B $+0.23$ \ci{-0.29}{+0.81}
attempts, Success@3 $103/21$, McNemar $p=3.5\times10^{-14}$; \ExactEIG{} B vs.\ \ELBOEIG{} B $-0.27$
\ci{-0.98}{+0.41}, $p=0.63$, Success@3 $23/23$. On the synthetic episodes the EIG-vs-random attempts contrast is
$-0.41$ \ci{-1.15}{+0.34} and Success@3 $90/23$ ($p=1.5\times10^{-10}$): the sign of the mean flips because the
random student dead-ends nine times in one vocabulary, while Success@3 tells the same story on both sets.
\textbf{Scratchpad diagnostics} (test split, turns $\ge2$): candidates equal to the exact consistent set
$0.723$ / $1.000$ / $0.843$ (\ELBOEIG{} / \ELBORand{} / \ExactEIG{}); secret retained $0.792$ / $1.000$ /
$0.897$; feedback line correct $0.997$ / $1.000$ / $0.995$; check verdicts correct $0.993$ / $1.000$ / $0.992$; exact-EIG-optimal rate over the counted attempts $0.613$ /
$0.471$ / $0.553$, against $1.000$ for the two EIG teachers on their training episodes (Figure~\ref{fig:inherit}). \textbf{Unrecoverable episodes.} \ELBOEIG{}: $9$ of $400$; in seven the secret is dropped from the
written list at turn two or three (a wrong verdict or a copying error between the check and candidates lines)
and one legal guess is repeated for the rest of the budget; one is a read-out slip with a correct one-word list;
one is a run of invalid attempts. \ExactEIG{}: $6$ of $400$ (four with the secret dropped, one read-out slip, one run of
letter-spaced invalid attempts after the list emptied). \ELBORand{}: none on the test split; nine on the
synthetic episodes, all in one vocabulary, all with exact bookkeeping and a repeated opener (a read-out failure
with a correct list). The teacher's traces never contain a wrong list, so the students have seen no example of a correction, and a
dropped secret is rarely restored: on the test split it re-enters the written list in one of the fifteen
episodes in which it is dropped, and three of those fifteen are solved. \textbf{Oracle read-out}: the secret is the only consistent word after at most three attempts
in $0.995$ / $0.892$ / $1.000$ of the \ELBOEIG{} / \ELBORand{} / \ExactEIG{} scratchpad students' episodes,
$0.960$ / $0.910$ / $0.935$ for the action-only students and $0.795$ for the base; of the action-only students'
$155$ / $79$ / $98$ unsolved episodes, $150$ / $74$ / $89$ had isolated the secret.
\textbf{Replicate runs.} The \ELBOEIG{} and \ExactEIG{} teachers choose identical guesses on every synthetic
training and validation episode, so under either target format their two students are trained on identical
dialogues with the identical recipe and seed; they differ only through non-deterministic GPU arithmetic. This is
the only replicate pair in the paper. With belief-scratchpad targets the replicates differ by $0.27$ attempts
\ci{-0.98}{+0.41} with identical Success@3 ($23/23$ discordant) and candidate-set accuracies of $0.72$ and
$0.84$; with action-only targets they differ by $3.79$ attempts \ci{+2.22}{+5.26} ($155$ against $98$ unsolved
episodes). The action-only \ELBOEIG{} student is also $3.93$ attempts \ci{+2.38}{+5.46} worse than the
action-only \ELBORand{} student, a gap of the same size as the replicate gap, while the second replicate is not
separated from the \ELBORand{} student ($+0.14$ \ci{-1.26}{+1.51}; Success@3 $p\ge0.28$ for all three pairs):
among students that fail at the read-out, differences of a few attempts are within run-to-run variation, and we
draw no conclusion about the query rule from them.

\subsection{Example Interactions}
\label{app:examples}
Movie Recommendation, belief-scratchpad target of the \ELBOEIG{} teacher (training episode; the dialogue of
Figure~\ref{fig:overview}; abridged): the first turn writes the prior
\texttt{<belief>0.45, ..., 0.45</belief>} and asks about the pair with the largest predicted disagreement; after
the answer ``No'' the belief moves to \texttt{0.48, 0.42, 0.45, 0.44, 0.44, 0.45, 0.59, 0.32}; the final turn
writes the belief and forty scores, and recommends the argmax. Word Guess, belief-scratchpad student: see the
target format in Appendix~\ref{app:scratchpads}, which the student reproduces with the exact set on $72\%$ of
turns. Circuit Decoding, action-only student: eighteen valid distinct queries and a $24$-bit guess in every
episode.

\section{Cross-Task Generalization}
\label{app:crosstask}

\subsection{Research Questions and Pre-Specified Decision Rules}
Two questions were fixed before any cross-task student was trained. Q1: does the information-seeking
\emph{policy} distilled on two task families transfer zero-shot to the third (judged by reference-EIG regret,
identification rate and task success on the held-out task, paired against the base model on identical
episodes)? Q2: does the information-gain \emph{criterion} transfer, that is, can a student trained to rank
candidate queries by their EIG on two tasks rank candidates on the third better than a shuffled-target control
and the base model? The decision rule for Q2 required a positive result in at least two of the three
directions, with confidence intervals excluding zero against both controls.

\subsection{Leave-One-Task-Out Protocol}
For each held-out task a student is trained on $1{,}800$ episodes from each of the two other tasks (frozen
subsets of the training banks; $128/840/832$ circuit episodes per inventory, $10$ users of each of $180$ score
tables, $20$ secrets of each of $90$ vocabularies) with the recipe of Section~\ref{sec:training}, using the
belief-scratchpad targets for movie and word and the action-only targets for circuit, and a loss that weights
the two source tasks equally per micro-batch. No data from the held-out task is read during training. Source-task
checks evaluate every student on both of its source tasks.

\subsection{Native and Unified Action Interfaces}
A format-compatibility pilot evaluated each in-domain student on the two other tasks under those tasks'
native prompts. The rule fixed in advance required the student's action to be parsable on at least $90\%$ of
turns for every student--task pair; four of six pairs passed ($0.946$--$1.000$) and the word student failed on
circuit ($0.772$) and movie ($0.289$). To remove format as a confound, all cross-task results below use a
\emph{unified interface}: a canonical outer wrapper \texttt{<action>...</action>} around the task's native
action, with a matching prompt for every task, under which every cross-task student emits a well-formed action
on at least $99\%$ of turns. The base model under the canonical prompts truncates its final circuit guess in
$49\%$ of episodes at the frozen final-token budget; this affects only the base's read-out metrics, not its
policy metrics, and no budget was changed after seeing results.

\subsection{Zero-Shot Transfer of the Information-Seeking Policy}
\begin{center}\footnotesize
\setlength{\tabcolsep}{4pt}
\begin{tabular}{llccc}
\toprule
Held-out task & Metric & Base & Cross-task student & In-domain student \\
\midrule
\multirow{3}{*}{Circuit ($300$)} & Success@18 $\uparrow$ & $0.000$ & $0.000$ & $0.920$ \\
 & EIG regret (bits) $\downarrow$ & $0.325$ & $0.540$ & $0.014$ \\
 & identified@18 $\uparrow$ & $0.457$ & $0.047$ & $0.997$ \\
\midrule
\multirow{3}{*}{Word ($400$)} & attempts $\downarrow$ & $30.55$ & $29.96$ & $3.37$ \\
 & Success@3 $\uparrow$ & $0.092$ & $0.075$ & $0.917$ \\
 & EIG regret (nats) $\downarrow$ & $0.104$ & $0.118$ & $0.167$ \\
\midrule
\multirow{3}{*}{Movie ($1{,}000$)} & Success@10 $\uparrow$ & $0.022$ & $0.028$ & $0.345$ \\
 & normalized rank $\downarrow$ & $0.595$ & $0.518$ & $0.100$ \\
 & EIG regret (nats) $\downarrow$ & $0.248$ & $0.240$ & $0.162$ \\
\bottomrule
\end{tabular}
\end{center}
Paired against the base on identical episodes: circuit regret $+0.215$ \ci{+0.196}{+0.235} (worse), identified
$-0.410$ \ci{-0.470}{-0.350}; word attempts $-0.59$ \ci{-2.13}{+0.93} ($p=0.51$), regret $+0.013$ (worse);
movie Success@10 $+0.006$ \ci{-0.006}{+0.018}, regret $-0.008$ \ci{-0.013}{-0.004} ($3\%$ of the base's). The
cross-task students repeat actions on $39\%$ (circuit) and $50\%$ (word) of turns once the history exceeds the
source tasks' horizon, and have no read-out for the held-out task. Source-task checks: on both of its sources every cross-task student is within $18$ points of the in-domain
student on Success@18 (circuit $0.797$ and $0.747$) and Success@10 (movie $0.307$ and $0.292$), and needs
$2.87$ and $4.48$ attempts on Word Guess against the in-domain student's $3.37$, so the held-out results are
interpretable.

\subsection{Transfer of the Information-Gain Criterion: Candidate-Ranking Formulation}
Each ranking example presents a state (history under the unified interface) and eight legal candidate actions
labeled with random letters; the target is the softmax of the candidates' reference EIG at temperature
$\tau$ (\emph{soft targets}, primary), the argmax (\emph{hard targets}, variant), or the soft targets permuted
within the candidate set (\emph{shuffled}, the decisive control with identical compute). Students are trained
jointly on the two source tasks' policy dialogues and ranking examples. Held-out banks are built from the test
episodes only: $4{,}800$ (circuit), $6{,}138$ (word) and $16{,}000$ (movie) examples from EIG-greedy and random
rollouts, of which $3{,}660$ / $1{,}746$ / $16{,}000$ have non-trivial EIG spread. Metrics on these contrast
examples, for a state with candidate EIGs $g_1,\dots,g_8$ under the reference posterior, extremes
$g_{\max}>g_{\min}$, the student's scores $z_1,\dots,z_8$ (the logits of the eight candidate letters at the
answer position) and its choice $\hat k=\arg\max_k z_k$:
\begin{equation}
\label{eq:rank-metrics}
\begin{aligned}
\text{normalized EIG regret}&=\frac{g_{\max}-g_{\hat k}}{g_{\max}-g_{\min}}\ \ (\text{primary}),\qquad
\text{top-1}=\mathbf 1\big[g_{\hat k}\ge g_{\max}-\epsilon\big],\\
\text{NDCG@8}&=\frac{\sum_{r=1}^{8}\tilde g_{\pi(r)}/\log_2(r+1)}{\sum_{r=1}^{8}\tilde g_{(r)}/\log_2(r+1)},\qquad
\tilde g_k=\frac{g_k-g_{\min}}{g_{\max}-g_{\min}},
\end{aligned}
\end{equation}
where $\pi$ orders the candidates by decreasing $z$, $\tilde g_{(r)}$ is the $r$-th largest normalized EIG and
$\epsilon$ is the task's tie tolerance of Appendix~\ref{app:metric-defs}, so top-1 counts a tied optimum as
correct. A uniform-random choice has expected normalized regret $\frac18\sum_k(g_{\max}-g_k)/(g_{\max}-g_{\min})$,
the reference line quoted with the results. Confidence intervals come from an episode-clustered paired
bootstrap with $10{,}000$ resamples.

\subsection{Results}
\begin{center}\footnotesize
\setlength{\tabcolsep}{4pt}
\begin{tabular}{lcccccc}
\toprule
Held-out task (contrast examples) & Base & Shuffled & Soft & Hard & Policy student & In-domain \\
\midrule
Circuit ($3{,}660$) & $0.457$ & $0.418$ & $0.509$ & $0.410$ & $0.500$ & $0.337$ \\
Word ($1{,}746$) & $0.385$ & $0.450$ & $0.400$ & -- & $0.398$ & $0.421$ \\
Movie ($16{,}000$) & $0.457$ & $0.448$ & $0.458$ & -- & $0.445$ & $0.437$ \\
\bottomrule
\end{tabular}
\end{center}
Normalized EIG regret (lower is better; uniform random choice scores $0.42$--$0.45$). Paired differences,
soft-target student minus base and minus shuffled control: circuit $+0.052$ \ci{+0.033}{+0.071} and $+0.091$
\ci{+0.068}{+0.114} (worse); word $+0.014$ \ci{-0.005}{+0.033} and $-0.050$ \ci{-0.072}{-0.029}; movie $+0.002$
\ci{-0.005}{+0.009} and $+0.011$ \ci{+0.003}{+0.018}.
Source checks pass in every direction (soft-target students reach
regret $0.001$--$0.235$ on their own sources against $0.395$--$0.474$ for the base). Robustness variants (permuted letters, task name removed, history removed, anonymized candidates), run for
the base model and the soft-target student in each direction, reveal no evidence of a shortcut: permuting the
candidate letters or removing the task name changes regret by at most $0.02$; removing the history changes it
by at most $0.03$ except for the circuit soft-target student ($0.509\to0.634$), which therefore relies on the
history; and anonymized payloads bring both models to the uniform-random level (base / soft-target student
against uniform random: circuit $0.421$ / $0.423$ vs.\ $0.419$; word $0.448$ / $0.461$ vs.\ $0.442$; movie
$0.446$ / $0.449$ vs.\ $0.446$). Zero of three directions meet the pre-specified rule.

\subsection{Analysis of Negative Transfer on Circuit Decoding}
Circuit Decoding is the only held-out cell with real dynamic range (the in-domain student is $0.120$ below the
base), and there transfer is negative: the soft-target student appears to carry a content-based preference for dense,
novel inputs that suits movie and word openings but not circuit ones. At the first turn its chosen query sets
$2.47$ of the three input bits on average, against $1.40$ for the highest-EIG candidate and $1.52$ for a random
one, with regret $0.59$ against $0.37$ for the base, whereas the hard-target and shuffled students choose
inputs of random density ($1.55$ and $1.52$ ones) and beat the base at the first turn (regret $0.29$ and $0.30$). On word the base already ranks well and even the in-domain student is
below it; on movie every condition is within $0.02$ of the base.

\subsection{Deviations from the Pre-Specified Plan}
(i) The switch from native to unified interfaces followed the failed compatibility pilot and was made before
any held-out result under the unified interface was seen. (ii) The joint policy-plus-ranking objective
replaced a ranking-only objective after a letter-only pilot never left the content-blind marginal; the change
was decided before any held-out ranking result of the joint objective was seen. (iii) One seed per condition.

\subsection{Discussion}
Neither the policy nor its criterion transfers zero-shot across these three task families under two-source
training. The result is consistent with the students learning task-specific procedures (which the in-domain
results show they learn well) rather than a portable notion of information value; whether broader task
distributions or a shared belief language would change this is open.

\section{Extended Related Work}
\label{app:related}
\textbf{Information seeking with language models.} Beyond the works cited in Section~\ref{sec:related},
falsification and rule-discovery benchmarks~\citep{jhaveri2026falsify} document a related failure, a
tendency not to seek disconfirming evidence (confirmation bias), which an EIG rule counteracts by construction. \textbf{Bayesian
experimental design with LMs.} Inference-time systems differ in what the LM supplies: a prior over hypotheses,
a likelihood, or the candidate queries; our teacher requires none of these from the LM and is therefore
verifiable, at the price of needing a task model. \textbf{Learning probabilistic reasoning from Bayesian
models.} \citet{qiu2025bayesianteaching} distill the predictions of a Bayesian model and report transfer to new
domains; our interactive setting adds the query rule and finds that, where the decision is an implicit
computation beyond the recorded interaction, the belief must be written for the decision to be learned, and
that transfer across our task families does not occur. \textbf{Distillation of
algorithms and traces.} Algorithm distillation~\citep{laskin2023algorithm} and search-trace
training~\citep{gandhi2024sos, lehnert2024beyondastar} teach transformers the dynamics of a procedure, and
progress-aligned supervision of looped models~\citep{wang2026par} shows that a final-answer target alone leaves
the intermediate state underdetermined; our belief scratchpad is the analogous exposure of a Bayesian filter's
state. The finding of Section~\ref{sec:res-components} that a more faithful teacher need not yield a better
student has a counterpart in model-based reinforcement learning, where \citet{kang2026drill} train world models
for the return of the policy learned inside them rather than for prediction error. \textbf{Amortized experimental
design.} Policy-based designers~\citep{foster2021dad, ivanova2021idad, blau2022rl} are trained by
maximizing an EIG-based objective; our students are trained by imitation of an explicit designer, which makes
the quality of the supervision measurable, and concurrent work~\citep{hartmann2026amortising, deng2026uncertainty} takes the
reward-based route with an LM policy. \textbf{Reinforcement learning for active reasoning.} The closest
reward-based work on belief tracking is that of \citet{zou2025belief}, who train Qwen2.5-7B-Instruct with
outcome rewards (PPO, GRPO, GSPO) and truncate a training trajectory once a task-specific proxy indicates that
the agent's belief has stopped improving, which stabilizes credit assignment. Three of their five tasks are
adapted from the benchmark used here, but they are not the released tasks, so their numbers cannot be set
against ours: Circuit Decoding becomes the identification of two hidden circuits from a list of ten candidate
structures within ten turns, on which Gemini~2.5~Pro reaches $92\%$ against $40\%$ on the released
task~\citep{badola2025mtp}, and Movie Recommendation uses five questions and an exact-match score on a
regenerated split; Word Guess is not included. The two routes are complementary: outcome reward with belief
control needs no model of the task, whereas a Bayesian teacher needs one and in return supplies verifiable
supervision for the query and for the belief.

\section{Reproducibility Details}
\label{app:repro}
\subsection{Software, Hardware and Runtime}
Teachers, generators and evaluators: Python 3.11, NumPy and PyTorch (CPU, teachers only). Training and
evaluation: PyTorch 2.8 (CUDA 12.8), Transformers 4.57, PEFT 0.17, bf16, SDPA attention, on NVIDIA A100
80\,GB GPUs; the \LLMteacher{} with vLLM 0.11. Runtimes: teacher rollouts on the synthetic episodes take under one CPU-hour per teacher on circuit and word and
about $4$ CPU-hours on movie ($3.3$ s per episode, Appendix~\ref{app:teacher-cost}; particle reference
$\approx66$ CPU-hours); training as in Appendix~\ref{app:training}; evaluation
$26$ s per episode for circuit students, $10$--$27$ s for action-only movie and word conditions, $120$ s per
episode for movie scratchpad students and $85$--$97$ s for word scratchpad students.
\subsection{Seeds and Determinism}
Training seed $0$ for every condition; teacher seeds $0$--$4$ paired by episode where the teacher is stochastic;
bootstrap seed $0$. Greedy decoding on the same GPU type reproduced every re-run episode bit-exactly: $300/300$ circuit episodes
for the base and action-only \ELBOEIG{} conditions evaluated twice, $100/100$ episodes in two independent
replications of word-guess conditions, and the smoke episodes regenerated inside every reused evaluation
(Appendix~\ref{app:protocol}). Training is not
bit-reproducible: the two word-guess students trained on identical dialogues (Appendix~\ref{app:word-extra})
end with validation losses of $0.0144$ and $0.0128$ (belief scratchpad) and $0.702$ and $0.777$ (action-only).

\subsection{Licenses of Models and Datasets}
Llama 3.1 models are used under the Llama 3.1 Community License; the benchmark is used under its release
license. The word pool is a public list of five-letter English words.

\section{Prompts}
\label{app:prompts}
\subsection{System and Turn Prompts per Task}
\textbf{Circuit Decoding, system prompt.}
\begin{verbatim}
You are playing Circuit Decoding, a multi-turn puzzle.

There are 3 hidden Boolean circuits, named A, B and C. Every circuit takes the
same 3 binary inputs (x0, x1, x2) and produces one output bit. The circuits are
built only from AND, OR and NOT gates. You are told only the total gate
inventory shared by the three circuits (how many AND, OR and NOT gates they use
in total); the wiring of each circuit is hidden.

Your goal is to recover the complete truth table of every circuit.

Rules:
1. On each turn you may ask exactly one query: the output of one circuit on one
   input assignment. Write it as <query>C(x0,x1,x2)</query>, where C is A, B or
   C and x0,x1,x2 are bits, for example <query>B(0,1,1)</query>.
2. The environment answers each query with a single bit, 0 or 1.
3. There are 24 possible queries (3 circuits x 8 input assignments). Each query
   may be asked at most once.
4. You have a budget of 18 queries. After the 18th answer you will be asked to
   submit your final answer.
5. On query turns respond with exactly one <query>...</query> tag and nothing else.
\end{verbatim}
\textbf{Circuit Decoding, final turn.}
\begin{verbatim}
Your query budget is exhausted. Submit your final answer now.

Give the complete truth tables of circuits A, B and C as one 24-bit string
inside <guess>...</guess>: bits 1-8 are the outputs of circuit A on inputs
(x0,x1,x2) = 000, 001, 010, 011, 100, 101, 110, 111 in that order, bits 9-16
are the outputs of circuit B on the same inputs, and bits 17-24 are the outputs
of circuit C. For example, <guess>000000000000000000000000</guess> would mean
that every circuit outputs 0 on every input. Respond with exactly one
<guess>...</guess> tag and nothing else.
\end{verbatim}
\textbf{Movie Recommendation, system prompt (action-only).}
\begin{verbatim}
You are playing Movie Recommendation, a multi-turn puzzle.

A user has a hidden preference function over 8 movie attributes:
score(movie) = w1*a1 + w2*a2 + ... + w8*a8, where a1..a8 are the movie's
attribute scores (given to you) and w1..w8 are the user's unknown non-negative
weights. You will be given two lists of movies with their attribute scores:
20 "seen" movies that you may ask the user about, and 40 "unseen" movies from
which you must finally recommend the one the user would score highest.

Rules:
1. On each turn ask exactly one question of the form <question>Would you prefer
   watching A over B?</question>, where A and B are two different movies from
   the seen list, written exactly as listed.
2. The user answers Yes, No or No Preference according to their hidden
   preference function.
3. You have a budget of 10 questions. After the 10th answer you will be asked
   for your recommendation.
4. On question turns respond with exactly one <question>...</question> tag and
   nothing else.
\end{verbatim}
\textbf{Movie Recommendation, final turn (action-only).}
\begin{verbatim}
Your question budget is exhausted. Recommend now: reply with exactly one
<recommend>TITLE</recommend> tag containing the title of one movie from the
unseen list, written exactly as listed, and nothing else.
\end{verbatim}
\textbf{Word Guess, first user message} (the vocabulary replaces \texttt{[VOCAB]}).
\begin{verbatim}
Let's play word guess. The secret word is from the following vocabulary: [VOCAB]
At each attempt, I will give feedback to you and specify which tiles are grey,
yellow and green. Green implies that the letter is in the word and in the
correct position, yellow implies that the letter is in the word but in the
wrong position and grey implies that the letter is not in the word.

For e.g., if the secret word is SPARK and you guessed PROOF, then the feedback
would be:
yellow,yellow,grey,grey,grey.

Your goal is to correctly guess the secret word in the minimum number of
attempts and will be evaluated on the number of attempts you take to arrive at
the secret word. Think step by step to ensure that you're taking the most
optimal actions to guess the secret word.

All of your guesses should be structured as follows:
<attempt>Your guess</attempt>

Make sure that all of your attempts are strictly from the vocabulary specified
above.
Start with your first attempt and make sure to follow the format specified above.
\end{verbatim}

\subsection{Action-Only and Belief-Scratchpad Templates}
The movie belief-scratchpad prompt replaces rule 4 of the action-only system prompt with:
\begin{verbatim}
4. Before every question, write your current estimate of the user's weights as
   <belief>w1, w2, w3, w4, w5, w6, w7, w8</belief> (eight numbers between 0.00
   and 0.90 with two decimals, in the attribute order given; before the first
   question use your prior estimate), then exactly one <question>...</question>
   tag, and nothing else. Update the estimate after each answer.
\end{verbatim}
and the final turn with:
\begin{verbatim}
Your question budget is exhausted. Recommend now: first write your final
estimate of the user's weights as <belief>w1, w2, w3, w4, w5, w6, w7, w8</belief>,
then <scores>TITLE: S; TITLE: S; ...</scores> listing every unseen movie in the
order given with S = its score under your estimated weights (two decimals), then
exactly one <recommend>TITLE</recommend> tag containing the title of the unseen
movie with the highest score, written exactly as listed, and nothing else.
\end{verbatim}
The word-guess scratchpad uses the same user prompt; the scratchpad lines are learned from the targets alone
(Appendix~\ref{app:scratchpads}).

\subsection{Unified-Interface Prompts for the Cross-Task Experiments}
The unified interface wraps each task's native action in \texttt{<action>...</action>} and adds one sentence to
each system prompt stating that every reply must consist of exactly one such tag containing the task's action;
scratchpad lines, where used, precede the tag. The full templates will be part of the code release upon
acceptance.

\end{document}